%% file: 00-main.tex
\documentclass[sigconf, prologue,dvipsnames,table, natbib=true,anonymous=false,review=false]{acmart}

\copyrightyear{2025}
\acmYear{2025}
\setcopyright{rightsretained}

\acmConference[SIGIR '25]{Proceedings of the 48th International ACM SIGIR Conference on Research and Development in Information Retrieval}{July 13--18, 2025}{Padua, Italy.}
\acmBooktitle{Proceedings of the 48th Int'l ACM SIGIR Conference on Research and Development in Information Retrieval (SIGIR '25), July 13--18, 2025, Padua, Italy}

\renewcommand\footnotetextcopyrightpermission[1]{}

\makeatother

\usepackage{enumitem}
\usepackage{longtable}
\newlist{inlinelist}{enumerate*}{1}
\setlist*[inlinelist,1]{%
  label=(\roman*),
}
\usepackage{subfigure}
\usepackage{xspace}
\usepackage{multirow}
\usepackage{adjustbox}
\usepackage{float}
\usepackage{color}
\usepackage{array}
\newcolumntype{H}{>{\setbox0=\hbox\bgroup}c<{\egroup}@{}}
\usepackage{booktabs}
\usepackage{microtype}
\usepackage{graphicx}
\usepackage{multicol}
\usepackage{algorithm}
\usepackage{algpseudocode}

\input{XX-notation}

\title{CIIR@LiveRAG 2025: Optimizing Multi-Agent Retrieval Augmented Generation through Self-Training}

\author{Alireza Salemi}
\affiliation{
\institution{University of Massachusetts Amherst}
\country{United States}
}
\email{asalemi@cs.umass.edu}

\author{Mukta Maddipatla}
\affiliation{\institution{University of Massachusetts Amherst}
\country{United States}
}
\email{mmaddipatla@umass.edu}

\author{Hamed Zamani}
\affiliation{
\institution{University of Massachusetts Amherst}
\country{United States}
}
\email{zamani@cs.umass.edu}

\begin{document}


\begin{abstract}
This paper presents \ourmethod, a multi-agent retrieval-augmented generation (RAG) framework composed of specialized agents for subtasks such as planning, searching, reasoning, and coordination. Our system uses a self-training paradigm with reward-guided trajectory sampling to optimize inter-agent collaboration and enhance response generation. Evaluated on DataMorgana-derived datasets during the SIGIR 2025 LiveRAG competition, \ourmethod outperforms conventional RAG baselines. We further analyze competition outcomes and showcase the framework’s strengths with case studies, demonstrating its efficacy for complex, real-world RAG tasks.
\end{abstract}






\maketitle

\section{Introduction}

RAG enhances large language models (LLMs) by integrating external retrieval mechanisms, addressing limitations like static knowledge and lack of grounding in current, verifiable sources \cite{kim2024retrievalenhancedmachinelearningsynthesis, erag, asai2023selfraglearningretrievegenerate}. Unlike traditional LLMs limited by fixed training data, RAG enables real-time access to relevant documents from sources such as search engines or databases \cite{ium, urag}, improving accuracy and relevance without retraining. Recently, autonomous agents have emerged as a powerful extension of RAG \cite{jin2025searchr1trainingllmsreason, singh2025agenticretrievalaugmentedgenerationsurvey}, capable of complex reasoning, tool use, and multi-step decision-making. Their modular design supports dynamic, goal-directed information synthesis, making them effective for real-world tasks requiring flexibility and depth.

While single-agent RAG systems are useful, they struggle with scalability, specialization, and effective context management. A single agent juggling multiple tasks—like query formulation, retrieval, synthesis, and validation—can suffer from context overlap, leading to inefficiencies and degraded performance. Multi-agent RAG systems address these issues through modular design. By assigning distinct roles (e.g., searcher, planner, summarizer, validator) to separate agents, each operates within a focused context, improving efficiency and task alignment. Inter-agent communication further enables better task decomposition and parallel execution, enhancing robustness and adaptability in complex reasoning workflows.

We introduce \ourmethod, a framework for building and optimizing multi-agent RAG systems. It consists of a set of task-specialized agents---such as a planner, searcher, and reasoner--each handling a distinct subtask. A central coordinator agent orchestrates the workflow by dynamically invoking agents, routing information, and monitoring progress toward a final response. The coordinator has control over the decision process and terminates the workflow once a satisfactory output is produced. To train agents, we adopt a self-training approach inspired by \citet{singh2024humandatascalingselftraining}, generating diverse agent interaction trajectories for each input. These are evaluated with a reward model, and high-reward trajectories are used as supervision. Agents are trained to reproduce these high-rewarded behaviors, promoting effective generation in future runs.

We present our experimental results from the LiveRAG\footnote{Competition website can be found at: \url{https://liverag.tii.ae/index.php}} competition, demonstrating that the proposed multi-agent framework outperforms the traditional retrieve-then-read RAG paradigm on a dataset derived from DataMorgana \cite{filice2025generatingdiverseqabenchmarks}, the competition’s data generation tool. Additionally, we analyze our performance on the competition test day and offer insights into the evaluation process. Finally, we include case studies highlighting examples where our multi-agent system achieves notably strong performance. We release our code to facilitate further research on this topic.\footnote{Available at: \url{https://github.com/muktac5/CIIR-LiveRAG}}

\section{Data Creation with DataMorgana}

To construct training and validation data for \ourmethod, we used DataMorgana \cite{filice2025generatingdiverseqabenchmarks} to generate a set of QA pairs. We defined 10 different question categories that were subsequently grouped into 5 combinations (definition of each category is reported in Appendix~\ref{app:datamoragana}):
\begin{itemize}[leftmargin=*]
    \item \textbf{User Expertise:} This category specifies the user's level of expertise on the topic by distinguishing between expert and novice users of the system (Figure~\ref{fig:user-category} in Appendix~\ref{app:datamoragana}).
    \item \textbf{Question Type:} This specifies the question phrasing style, varying along dimensions such as length (short vs. long), formality (natural vs. query-like), and verbosity (concise vs. verbose), resulting in six distinct combinations (Figure~\ref{fig:question-phrasing-category} in Appendix~\ref{app:datamoragana}).
    \item \textbf{Answer Type:} This defines the answer phrasing, varying along the dimensions of natural vs. formal and concise vs. verbose, resulting in four distinct combinations (Figure~\ref{fig:answer-category} in Appendix~\ref{app:datamoragana}).
    \item \textbf{Question Intent:} This defines the intention behind the question, encompassing categories such as clarification, opinion, comparison, yes/no, and hypothetical questions (Figure~\ref{fig:question-intent-category} in Appendix~\ref{app:datamoragana}).
    \item \textbf{Answer Intent:} This specifies intents behind responses, categorized as either factual or open-ended (Figure~\ref{fig:question-factuality-category} in Appendix~\ref{app:datamoragana}).
    \item \textbf{Premise Inclusion:} This indicates whether the question includes user-specific information, resulting in two variations: with premise and without premise (Figure~\ref{fig:question-premise-category} in Appendix~\ref{app:datamoragana}).
    \item \textbf{Lexical Similarity:} This specifies the lexical alignment between questions and documents: using similar terms, different terms, or referencing unpopular entities (Figure~\ref{fig:document-variation-category} in Appendix~\ref{app:datamoragana}).
    \item \textbf{Aspect Granularity:} This defines whether the question capture a single or multiple aspects of the topic (Figure~\ref{fig:question-aspect-category} in Appendix~\ref{app:datamoragana}).
    \item \textbf{Interaction Type:} This defines whether the question initiates a conversation or is a follow-up question (Figure~\ref{fig:question-turn-category} in Appendix~\ref{app:datamoragana}).
    \item \textbf{Document Granularity:} This specifies whether one or two documents are used to generate the question, using the feature in the DataMorgana platform (Figure~\ref{fig:document-category} in Appendix~\ref{app:datamoragana}).
\end{itemize}

After interacting with DataMorgana, we find that using all categories simultaneously does not yield high-quality outputs. Consequently, we grouped them into five combinations when invoking DataMorgana for data generation. All combinations include the following core dimensions: User Expertise, Question Type, Answer Type, Document Granularity, Interaction Type, and Aspect Granularity. The remaining categories in each combination are as follows:
\begin{itemize}[leftmargin=*]
    \item Answer Intent and Lexical Similarity.
    \item Premise Inclusion.
    \item Question Intent.
    \item Premise Inclusion and Question Intent.
    \item Lexical Similarity and Premise Inclusion.
\end{itemize}
where we generate 4,500 question-answer pairs with DataMorgana, allocating 3,500 for training and 1,000 for evaluation of our system. Details of the data generation process are provided in Appendix~\ref{app:datamoragana}.

\section{The \ourmethod Framework}

\begin{figure}
    \centering
    \includegraphics[width=0.95\linewidth]{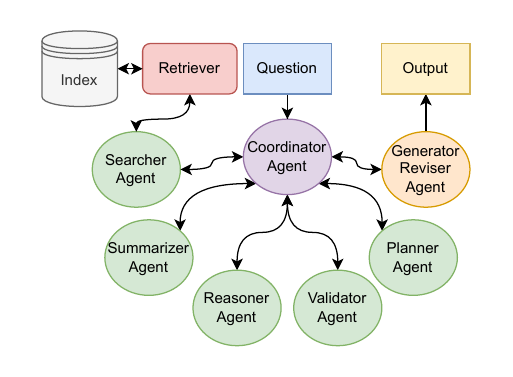}
    \vspace{-0.8cm}
    \caption{Overview of multi-agent RAG.}
    \label{fig:overview}
    \vspace{-0.6cm}
\end{figure}

Our framework employs six action-performing agents and one coordinator agent to generate responses. An overview of \ourmethod is shown in Figure~\ref{fig:overview}. This section provides implementation and training details of the multi-agent system.

\subsection{Agents in \ourmethod}
\label{sec:agents}

This section describes the implementation of the agents.

\subsubsection*{\textbf{Coordinator}}

This agent serves as the entry point of \ourmethod. It receives a question and a set of agents with defined input and output formats, and is responsible for assigning tasks to these agents based on their respective expertise. In each turn, the coordinator selects an appropriate agent based on the current state alongside the rational behind selection of this agent, formats and passes the necessary inputs, and waits for the agent to complete its task and return outputs in the expected format. Upon receiving the output, the coordinator updates its internal state by appending the new outputs to the conversation history, and updates either the response (if one was generated) or the set of supporting documents (if new information was retrieved). This process is repeated iteratively. Once the computation budget is exhausted or the coordinator determines that no further actions are required and a high-quality response has been produced, it terminates the process and returns the final response along with the supporting documents. This agent use instruct Qwen 2.5 with 7B parameters \cite{qwen2025qwen25technicalreport} with the prompt shown in Figure~\ref{fig:coordinator-agent} in Appendix~\ref{app:agents-prompts} in each step. The detailed implementation is provided in Algorithm~\ref{alg:coordinator} in Appendix~\ref{app:agents-prompts}.

\subsubsection*{\textbf{Searcher}}

To access the corpus and collect information necessary for answering the question, the coordinator can invoke the searcher agent by providing the question, the information gathered so far, and a suggested set of aspects to guide the search. The searcher agent begins by generating a search query, which is executed using the sparse Lion retrieval model with 1B parameters (detailed implementation and corpus processing described in Appendix~\ref{app:retrieval}) \cite{zeng2025scalingsparsedenseretrieval} to retrieve two documents. Next, the searcher agent evaluates the relevance of each retrieved document, providing justifications and marking the relevant ones. At this stage, the agent has three main options: (1) continue the search using the same query to retrieve the next two documents, (2) modify the search query and explain the rationale for the new query, or (3) terminate the search, explaining why, and return the relevant documents found so far. This process continues until either the maximum retrieval budget is reached or the agent decides that sufficient information has been gathered. This design allows the searcher agent to dynamically determine the appropriate amount of information to collect for each query. The agent uses the instruct Qwen 2.5 model with 7B parameters \cite{qwen2025qwen25technicalreport} and the prompt shown in Figure~\ref{fig:searcher-agent} in Appendix~\ref{app:agents-prompts} in each step. The detailed implementation of this agent is provided in Algorithm~\ref{alg:searcher}.

\subsubsection*{\textbf{Planner}}

This agent is responsible for generating a sequence of steps required to produce a response to the question, based on the given question and the information collected so far. Although it can be invoked at any stage by the coordinator agent, it is recommended to be called at the beginning of the response generation process. The prompt used for this agent is shown in Figure~\ref{fig:planner-agent} in Appendix~\ref{app:agents-prompts}. This agent uses an instruct Qwen 2.5 with 7B parameters \cite{qwen2025qwen25technicalreport}.

\subsubsection*{\textbf{Summarizer}}

As the conversation between agents grows longer, it becomes increasingly challenging for the coordinator to track all relevant details. To mitigate this, the coordinator can invoke this agent and provide selected information to summarize the conversation or the retrieved and collected content up to that point, using the prompt shown in Figure~\ref{fig:summarizer-agent} in Appendix~\ref{app:agents-prompts}. This agent uses an instruct Qwen 2.5 with 7B parameters \cite{qwen2025qwen25technicalreport}.

\subsubsection*{\textbf{Reasoner}}

When step-by-step reasoning or analysis is required regarding the retrieved information, actions taken, or any other aspect of the process, this agent can be invoked by providing the question, relevant information, and a specific aspect to reason about. This agent uses the prompt shown in Figure~\ref{fig:reasoner-agent} in Appendix~\ref{app:agents-prompts}. This agent uses an instruct Qwen 2.5 with 7B parameters \cite{qwen2025qwen25technicalreport}.

\subsubsection*{\textbf{Validator}}

Sometimes, questions may specify certain criteria that must be satisfied. This agent can be invoked when a response has been generated and the coordinator needs to ensure that all criteria are addressed. The coordinator provides the question, retrieved information, and a response to this agent, which then extracts the criteria from the question and verifies whether they are fulfilled in the response and provides rationals for each of its decisions. This process uses the prompt shown in Figure~\ref{fig:validator-agent} in Appendix~\ref{app:agents-prompts} using an instruct Qwen 2.5 with 7B parameters \cite{qwen2025qwen25technicalreport}.

\subsubsection*{\textbf{Generator/Reviser}}

To generate a response, the coordinator invokes this agent by providing the question, supporting information collected during the process, a response plan, and a set of key aspects that should be included. The agent then generates a response that incorporates these elements. Additionally, if the coordinator determines that the response requires revision---either due to newly collected information or unmet criteria---it can call this agent again with suggestions outlining the deficiencies and how to address them. This agent uses the prompts shown in Figure~\ref{fig:generator-reviser-agent} in Appendix~\ref{app:agents-prompts}. Due to competition constraints, we use the instruction-tuned Falcon 3 model with 10B parameters \cite{Falcon3} for this agent.

\subsection{Optimization through Self-Training}
\label{sec:optimization}

End-to-end training of multi-agent systems is challenging due to computational inefficiency and the difficulty of propagating gradients across agents \cite{gogineni2023efficientmultiagentlearningsystems, Gogineni2024CharacterizingAO}. To simplify the training process, we make a key assumption: agents operate independently and function solely based on their provided inputs. With this assumption, the probability of a trajectory of actions taken by the agents, denoted as $\tau = a_1 a_2 \ldots a_n$, simplifies from the full joint distribution
$$
p(a_1 \ldots a_n \mid x) = p(a_1 \mid x) \, p(a_2 \mid a_1, x) \, \ldots \, p(a_n \mid a_{n-1}, \ldots, a_1, x)
$$
to the following factorized form:
$$
p(\tau \mid x) = p(a_1 \ldots a_n \mid x) = p(a_1 \mid x) \, p(a_2 \mid x) \, \ldots \, p(a_n \mid x),
$$
assuming conditional independence of agent actions given the input $x$. With this simplification, we can use Self-Training \cite{singh2024humandatascalingselftraining} to optimize the system end to end. We first sample $T = 8$ diverse trajectories for each input in the training set by applying a high sampling temperature. A reward model $RM$ is then applied to the final response of each trajectory for scoring. We retain only the trajectory with the highest reward score for each input (allowing up to three trajectories in case of ties, to avoid overfitting on simpler examples). These selected trajectories are used to train the agents to reproduce the optimal sequence of actions. Following \citet{singh2024humandatascalingselftraining}, we normalize the reward by setting the reward $r(\tau, x)$ of the best-performing trajectory to 1 (by setting the threshold score to match the highest observed reward) and others to zero. Therefore, the training objective is to maximize:
\begin{align*}
\mathbb{E}_{\tau \sim D} 
\left[ r(\tau, x)\log{p(\tau \mid x)} \right] 
= 
\mathbb{E}_{\tau \sim D} 
\left[  \sum_{a_i \in A} \log{p(a_i \mid x)} \right]
\end{align*}
where $A$ is the set of actions performed by trainable agents (i.e., all agents except the generator/reviser that is fixed due to competition constraints) and $D$ is the set of all selected trajectories after applying reward model. This objective is equivalent to supervised fine-tuning \cite{seq2seq} of the agents on their own best-performing outputs within the selected trajectories. This encourages each agent to replicate the behavior observed in the highest-reward (i.e., most successful) trajectory, promoting consistency with effective sequences. The training details and parameters are provided in Appendix~\ref{app:train-details}.

\subsubsection*{\textbf{Reward Model}}

The competition does not provide a publicly available scoring function. However, based on the provided guidelines, we define and utilize two distinct reward signals (more implementation details are explained in Appendix~\ref{app:train-reward}):
\begin{itemize}[leftmargin=*]
    \item correctness: We adopt the recall-oriented nugget-based reward proposed by \citet{pradeep2025greatnuggetrecallautomating}. In this approach, we first extract the atomic aspects that should be present in the output from the ground truth response using the prompt shown in Figure~\ref{fig:correctness-extract} in Appendix~\ref{app:train-reward}. Each extracted aspect is then evaluated for alignment with the generated response using the prompt shown in Figure~\ref{fig:correctness-match} in Appendix~\ref{app:train-reward}, with scores assigned on a scale from –1 to 2. These scores are subsequently normalized to the range [0, 1]. The final correctness reward is computed as the average normalized score across all nugget aspects. The implementation is detailed in Algorithm~\ref{alg:correctness} in Appendix~\ref{app:train-reward}.
    
    
    \item Faithfulness: This reward model evaluates the faithfulness of the generated response with respect to the retrieved information. Following \citet{ragas}, we first extract the atomic aspects from the generated response using the prompt in Figure~\ref{fig:faithfulness-extract} in Appendix~\ref{app:train-reward}. Each aspect is then scored for how well it is supported by the retrieved documents using the prompt in Figure~\ref{fig:faithfulness-match} in Appendix~\ref{app:train-reward}, on a scale from –1 to 1, which is subsequently normalized to the range [0, 1]. The final faithfulness score is computed as the average normalized score across all extracted aspects. The implementation is detailed in Algorithm~\ref{alg:faithfulness} in Appendix~\ref{app:train-reward}.
\end{itemize}

To compute the final reward, we run each reward model five times and take the average score for each. We observed that \ourmethod performs worst on the correctness reward. To emphasize improvement in this area, we assign a weight of 4 to the correctness reward and a weight of 1 to faithfulness reward, and compute a weighted average to obtain the final reward. We use instruct Qwen 2.5 with 14B parameters \cite{qwen2025qwen25technicalreport} as the LLM for all reward functions.

\section{Experiments}

This section presents our findings. The experimental setup is described in Appendix~\ref{app:inference-setup}. For comparison, we consider two baseline models: a standard language model without retrieval augmentation (vanilla LLM) and a retrieval-augmented generation model (vanilla RAG) that follows a retrieve-then-read paradigm. Detailed implementation of the baseline systems is provided in Appendix~\ref{app:baselines}.

\begin{figure}
    \centering
    \includegraphics[width=\linewidth]{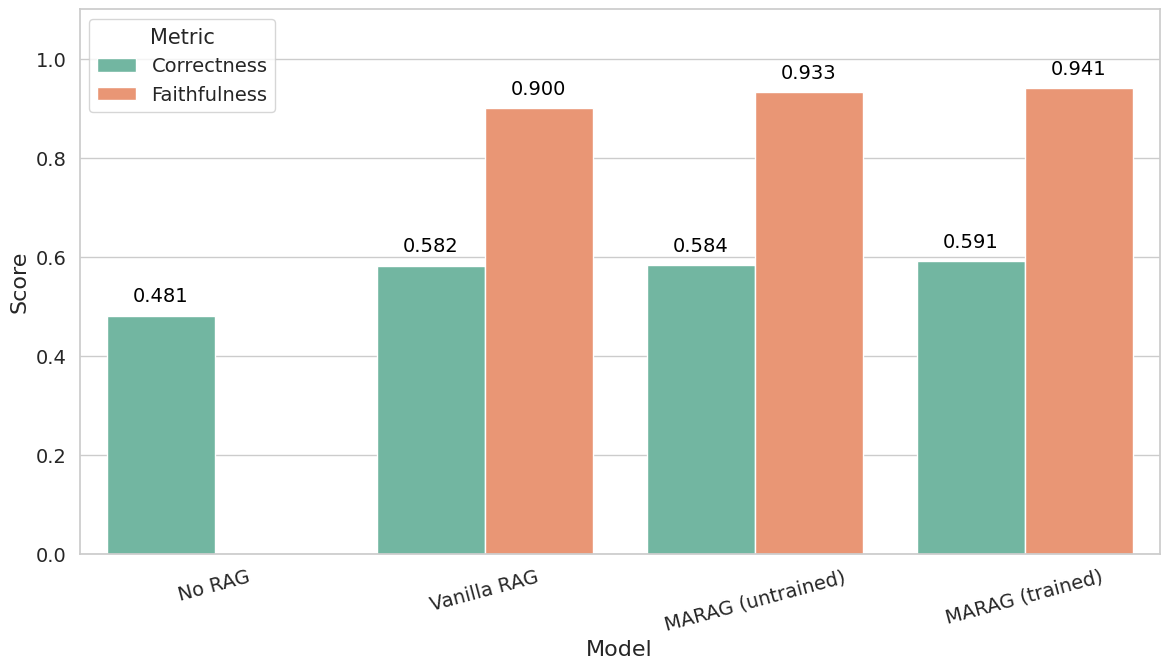}
    \vspace{-0.6cm}
    \caption{Evaluation of \ourmethod and baselines using the introduced reward models in Section~\ref{sec:optimization} and Appendix~\ref{app:train-reward}.}
    \label{fig:results-our-test}
    \vspace{-0.4cm}
\end{figure}

\subsubsection*{\textbf{Results on the created test set from DataMorgana}}

We compare \ourmethod with two baseline approaches: a vanilla LLM without retrieval augmentation (RAG) and a vanilla RAG system employing a retrieve-then-read paradigm. Evaluation results on the test set derived from DataMorgana are presented in Figure~\ref{fig:results-our-test}. The findings indicate that \ourmethod, when optimized using the method described in Section~\ref{sec:optimization}, outperforms both baselines in terms of correctness and faithfulness. Notably, even without training, \ourmethod achieves comparable correctness to the vanilla RAG while demonstrating superior faithfulness. Across all configurations, RAG-based methods---\ourmethod and vanilla RAG---consistently outperform the non-RAG baseline. Based on these results, \ourmethod demonstrates the best overall performance on the constructed dataset from DataMorgana.

\subsubsection*{\textbf{Results on the test set from the competition}}

\begin{table}
    \caption{Results of the top 20 participating teams in the LiveRAG competition. The competition-defined evaluation metrics assign scores in the range of -1 and 2 for correctness, and -1 and 1 for faithfulness. The final ranking is determined by sorting the teams based on correctness scores.}
    \vspace{-0.2cm}
    \centering
    \begin{tabular}{ll|cc}
    \toprule
    Rank & Team & Correctness & Faithfulness \\
    \midrule
    1 & Magikarp & 1.231 & 0.656 \\
    2 & UDInfo & 1.200 & 0.623 \\
    3 & RMIT-ADMS & 1.199 & 0.477 \\
    4 & RAGtifier & 1.134 &	0.552 \\
    5 & HLTCOE & 1.070 & 0.340 \\
    6 & Ragmatazz & 1.011 & 0.519 \\
    \midrule
    7 & \ourmethod & 0.996 & 0.418 \\
    \midrule
    8 & RUC\_DeepSearch & 0.969 & 0.387 \\
    9 & Ped100X	& 0.928 & 0.043 \\
    10 & PRMAS-DRCA & 0.922 & 0.410 \\
    11 & Emorag & 0.890 & 0.556 \\
    12 & Graph-Enhanced RAG & 0.875 & 0.529 \\
    13 & Hybrid Search with Graph & 0.875 & 0.315 \\
    14 & Multi-Agent Adaptive RAG & 0.836 & 0.200 \\
    15 & Starlight & 0.818 & 0.433 \\
    16 & BagBag & 0.694 & -0.911 \\
    17 & UniClustRAG & 0.685 & 0.460 \\
    18 & METURAG & 0.673 & 0.325 \\
    19 & NoobRAG & 0.655 & 0.154 \\
    20 & DeepRAG & 0.566 & 0.097 \\
    \bottomrule
    \end{tabular}
    \label{tab:results-competition}
    \vspace{-0.4cm}
\end{table}

The results of \ourmethod and other top 20 participating systems on the test set provided by the LiveRAG competition prganizers during the official test day are presented in Table~\ref{tab:results-competition}. The specific implementation details of the correctness and faithfulness metrics reported in Table~\ref{tab:results-competition} are not publicly available. However, the correctness scores range from -1 to 2, while the faithfulness scores range from -1 to 1, as defined by the competition organizers. We report the raw, unnormalized scores as provided by the competition evaluation process. According to these results, which are ranked based on the correctness metric, \ourmethod achieves the 7th position among all participating teams.

\subsubsection*{\textbf{Unstability of the recall-oriented nugget-based evaluation of correctness}}

In one of our experiments, we found that simply prompting the LLM to produce longer responses---without any additional training---raised the nugget-based recall reward from 0.528 to 0.584. This highlights a key limitation in current evaluation methods: as responses become longer, they are more likely to include a greater number of nuggets, thereby inflating recall-oriented rewards. However, this comes at the cost of conciseness and clarity, resulting in redundant and potentially less helpful output for users. This suggests that recall-focused metrics may encourage verbosity over precision. To address this, future evaluation strategies should aim to balance both recall and precision by incorporating metrics that also account for relevance and brevity, better aligning with user expectations for clear and informative responses.

\section{Case Study}

To show the effectiveness of \ourmethod, we present two case studies that illustrate the system's decision-making trajectories and inter-agent interactions in response to user queries. Detailed descriptions of these case studies are provided in Appendix~\ref{app:case-study}. We examine two notable behaviors exhibited by the system after training, as observed in its responses to the queries: “safety concerns hydrogen steam reforming” and “How did John Ball influence golf history at Hoylake, and what strategic challenges does the course present regarding out of bounds?”. These two notable behaviors emerged:
\begin{itemize}[leftmargin=*]
    \item \textbf{Breaking down the question and collecting information about each aspect:} In both cases, as illustrated in Table~\ref{case-study-1} and Table~\ref{case-study-2} in Appendix~\ref{app:case-study}, the system decomposes each question into multiple aspects that must be addressed to formulate a comprehensive response. The coordinator invokes the searcher agent twice in each case to retrieve information corresponding to two distinct facets of the query. The retrieved information is then relayed to other agents responsible for response generation, reasoning, and validation. These interactions demonstrate that the coordinator has effectively learned how to orchestrate the contributions of other agents—particularly the searcher, which plays a critical role in supporting information acquisition.

    \item \textbf{Changing search query after retrieval failure:} In both examples presented in Table~\ref{case-study-1} and Table~\ref{case-study-2} in Appendix~\ref{app:case-study}, the searcher agent demonstrates adaptive behavior in response to retrieval failures. When the initial query fails to yield sufficient or relevant results, the searcher does not immediately reformulate the query. Instead, it reuses the same query for up to five retrieval attempts, aiming to extract useful information from lower-ranked documents---accounting for potential imperfections in the retriever. Only after these repeated attempts does the agent revise the query to explore alternative formulations. This behavior indicates that the searcher has learned a robust strategy for balancing persistence with query reformulation in the presence of retrieval noise or deficiencies.
\end{itemize}

Further details of the case studies are provided in Appendix~\ref{app:case-study}, which highlight several notable emergent behaviors exhibited by the system after training.

\section{Conclusion}

This paper presented \ourmethod, a multi-agent framework designed for RAG, along with a training methodology to optimize inter-agent communication. We trained \ourmethod using data generated from DataMorgana and participated in the SIGIR 2025 LiveRAG competition. Experimental results demonstrated that \ourmethod consistently outperforms standard vanilla RAG baselines. Additionally, \ourmethod achieved a 7th place ranking based on automatic evaluation in the competition, highlighting its effectiveness in real-world settings.

\bibliographystyle{ACM-Reference-Format}
\bibliography{XX-references}

\appendix

\section{Details of Creating Data with DataMorgana}
\label{app:datamoragana}
\begin{figure}
    \centering
    \includegraphics[width=1.0\linewidth]{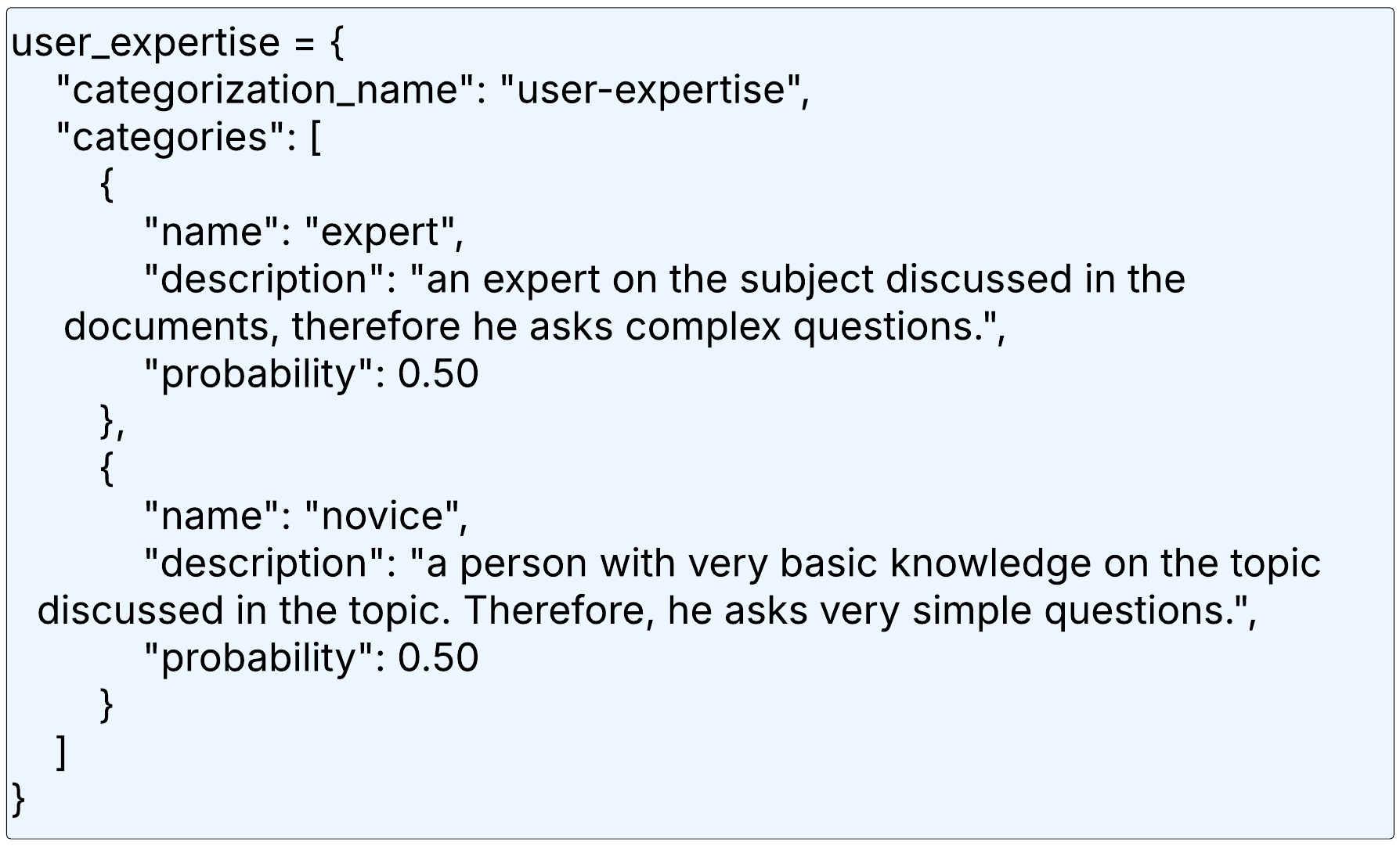}
    \caption{User expertise categorization schema and probabilities.}
    \label{fig:user-category}
\end{figure}

\begin{figure}
    \centering
    \includegraphics[width=1.0\linewidth]{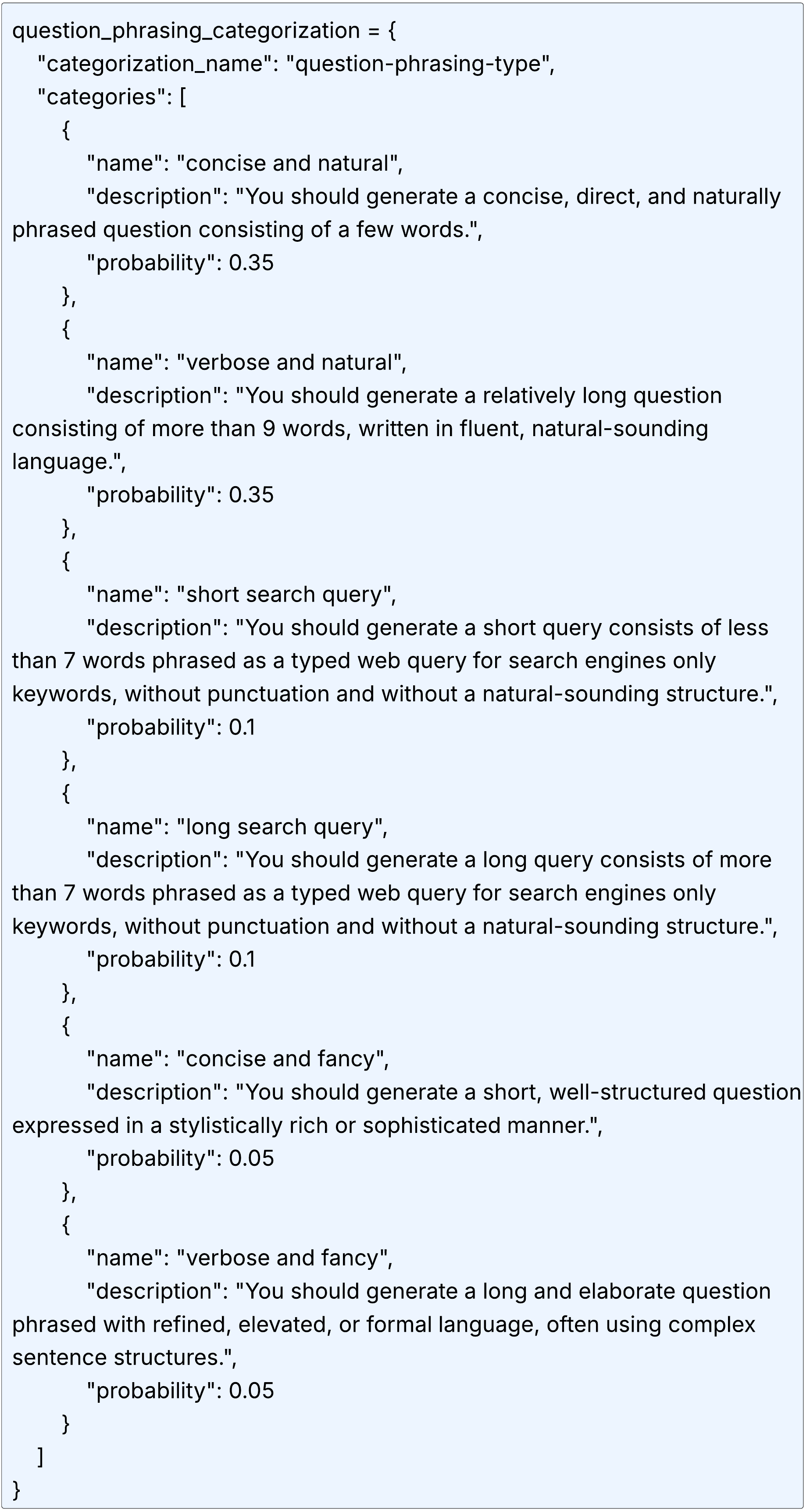}
    \caption{Question type categorization schema and probabilities.}
    \label{fig:question-phrasing-category}
\end{figure}

\begin{figure}
    \centering
    \includegraphics[width=1.0\linewidth]{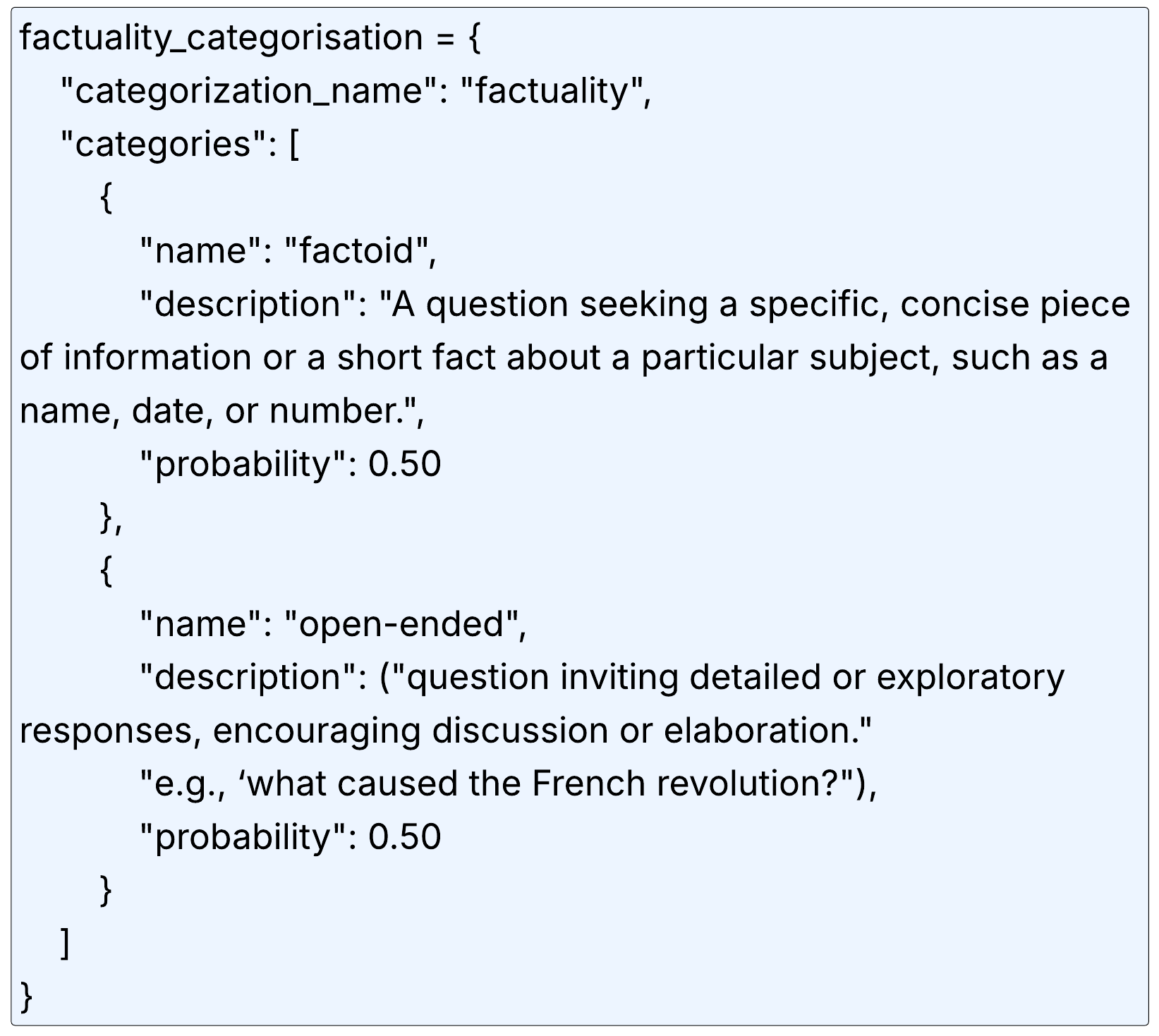}
    \caption{Answer Intent categorization schema and probabilities.}
    \label{fig:question-factuality-category}
\end{figure}

\begin{figure}
    \centering
    \includegraphics[width=1.0\linewidth]{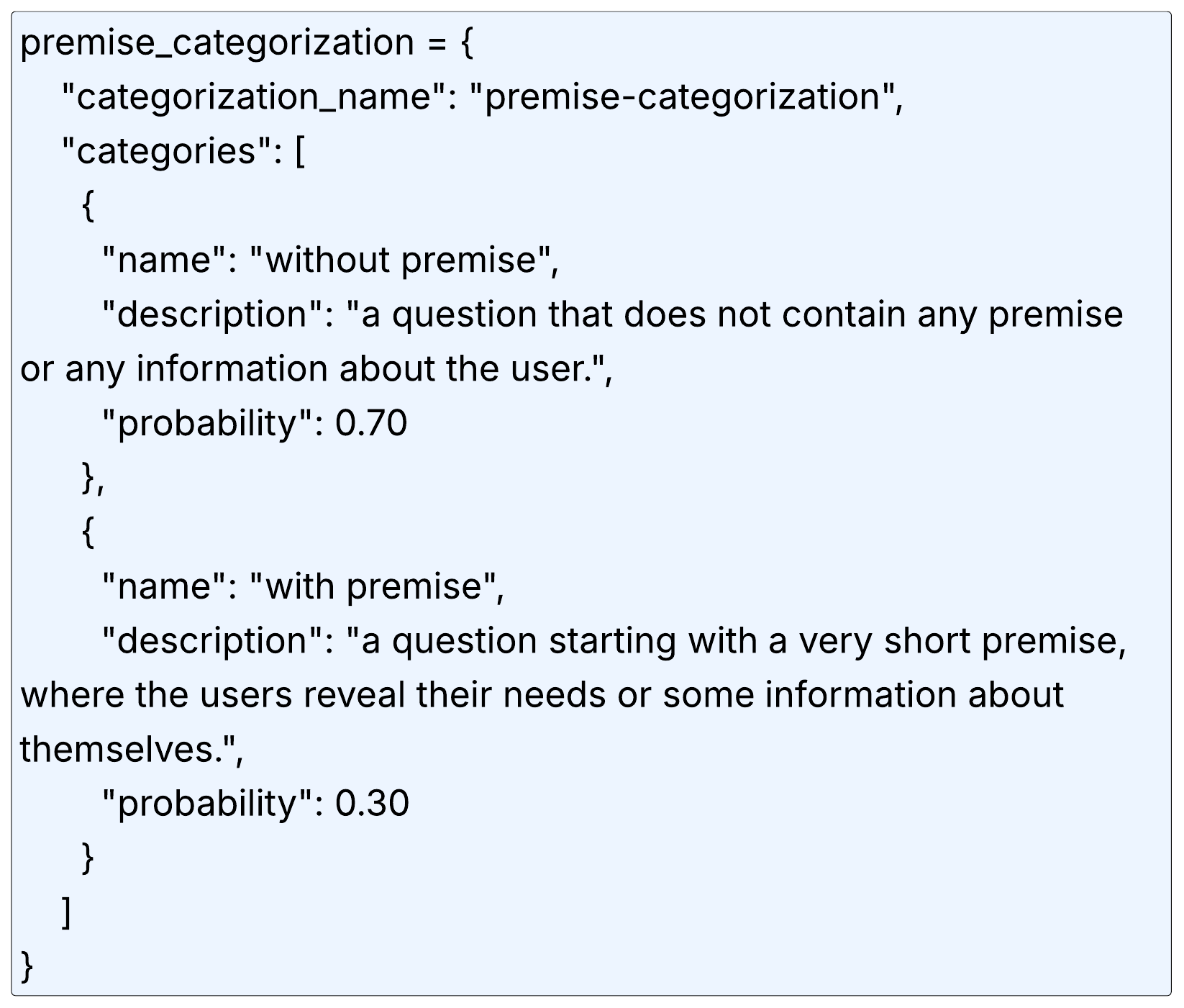}
    \caption{Premise Inclusion categorization schema and probabilities.}
    \label{fig:question-premise-category}
\end{figure}

\begin{figure}
    \centering
    \includegraphics[width=1.0\linewidth]{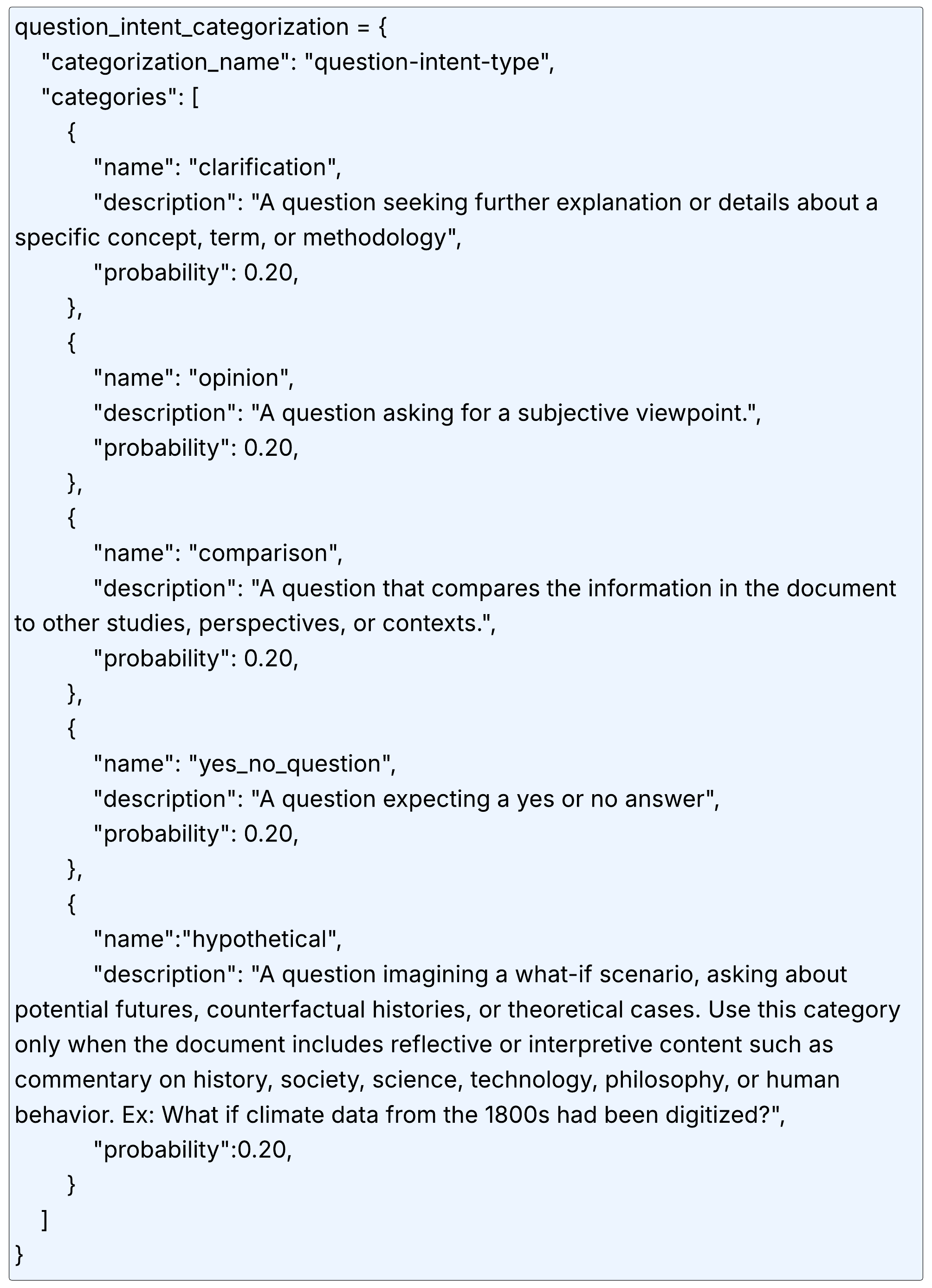}
    \caption{Question Intent categorization schema and probabilities.}
    \label{fig:question-intent-category}
\end{figure}

\begin{figure}
    \centering
    \includegraphics[width=1.0\linewidth]{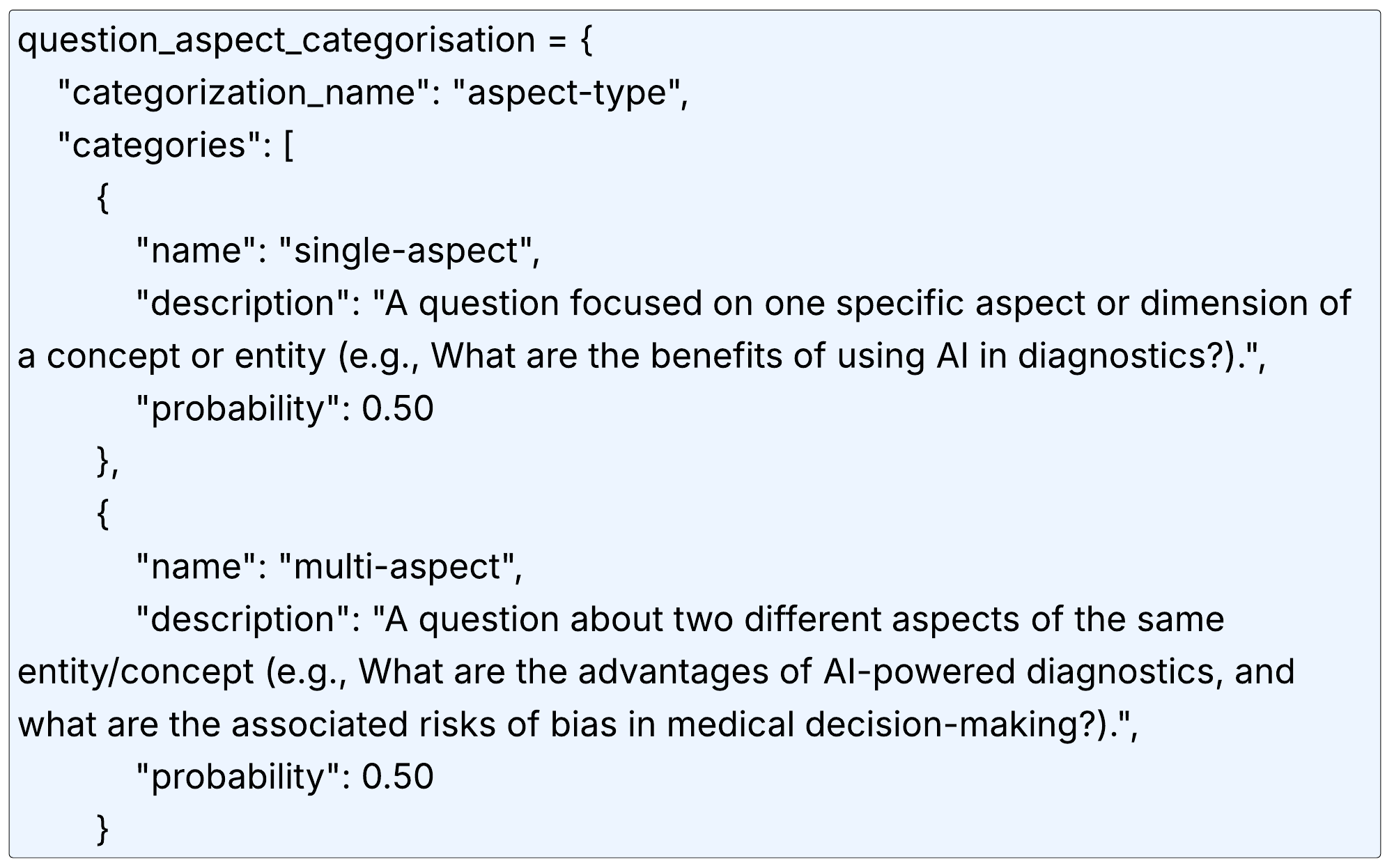}
    \caption{Aspect Granularity categorization schema and probabilities.}
    \label{fig:question-aspect-category}
\end{figure}

\begin{figure}
    \centering
    \includegraphics[width=1.0\linewidth]{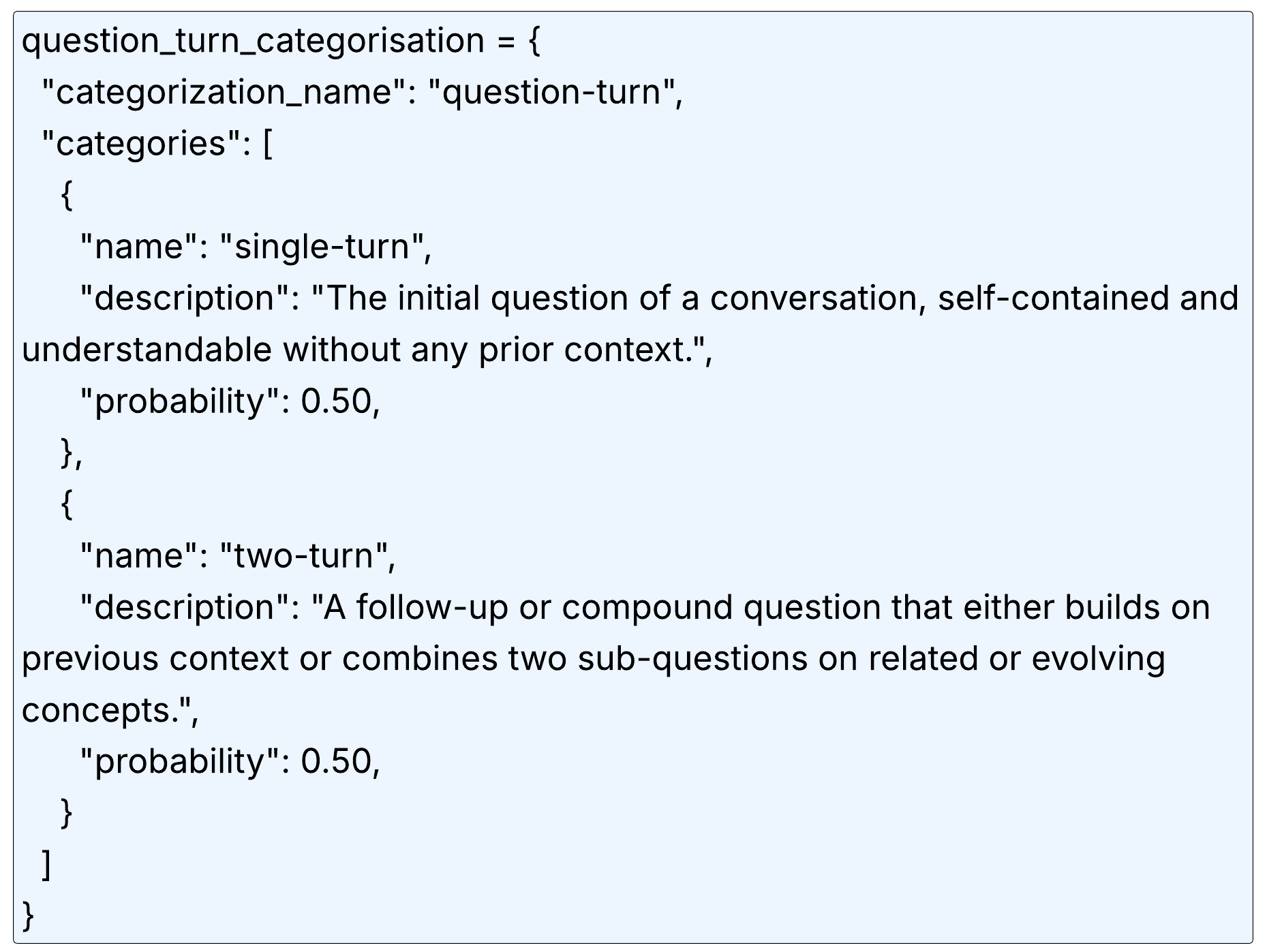}
    \caption{Interaction type categorization schema and probabilities.}
    \label{fig:question-turn-category}
\end{figure}

\begin{figure}
    \centering
    \includegraphics[width=1.0\linewidth]{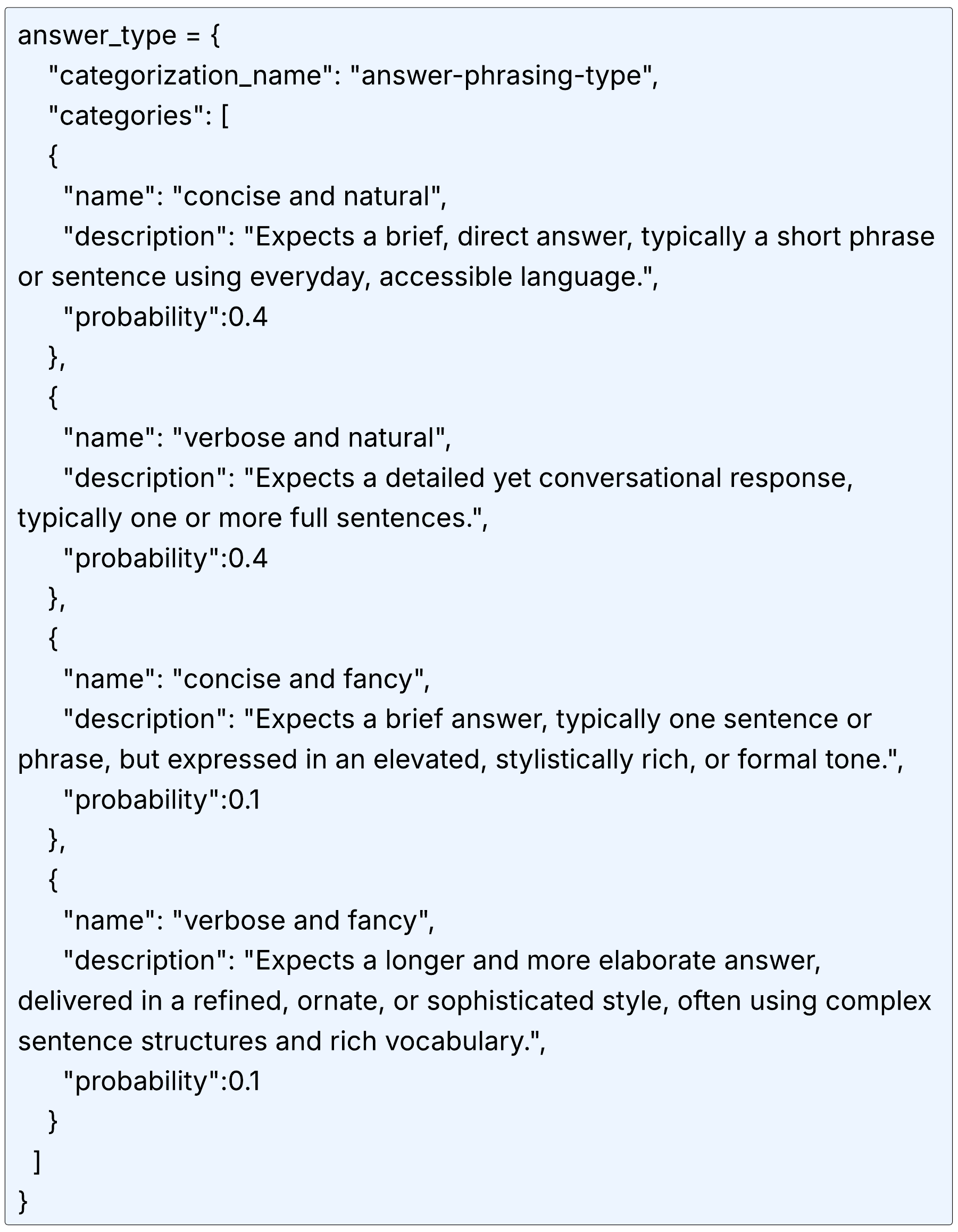}
    \caption{Answer type categorization schema and probabilities.}
    \label{fig:answer-category}
\end{figure}

\begin{figure}
    \centering
    \includegraphics[width=1.0\linewidth]{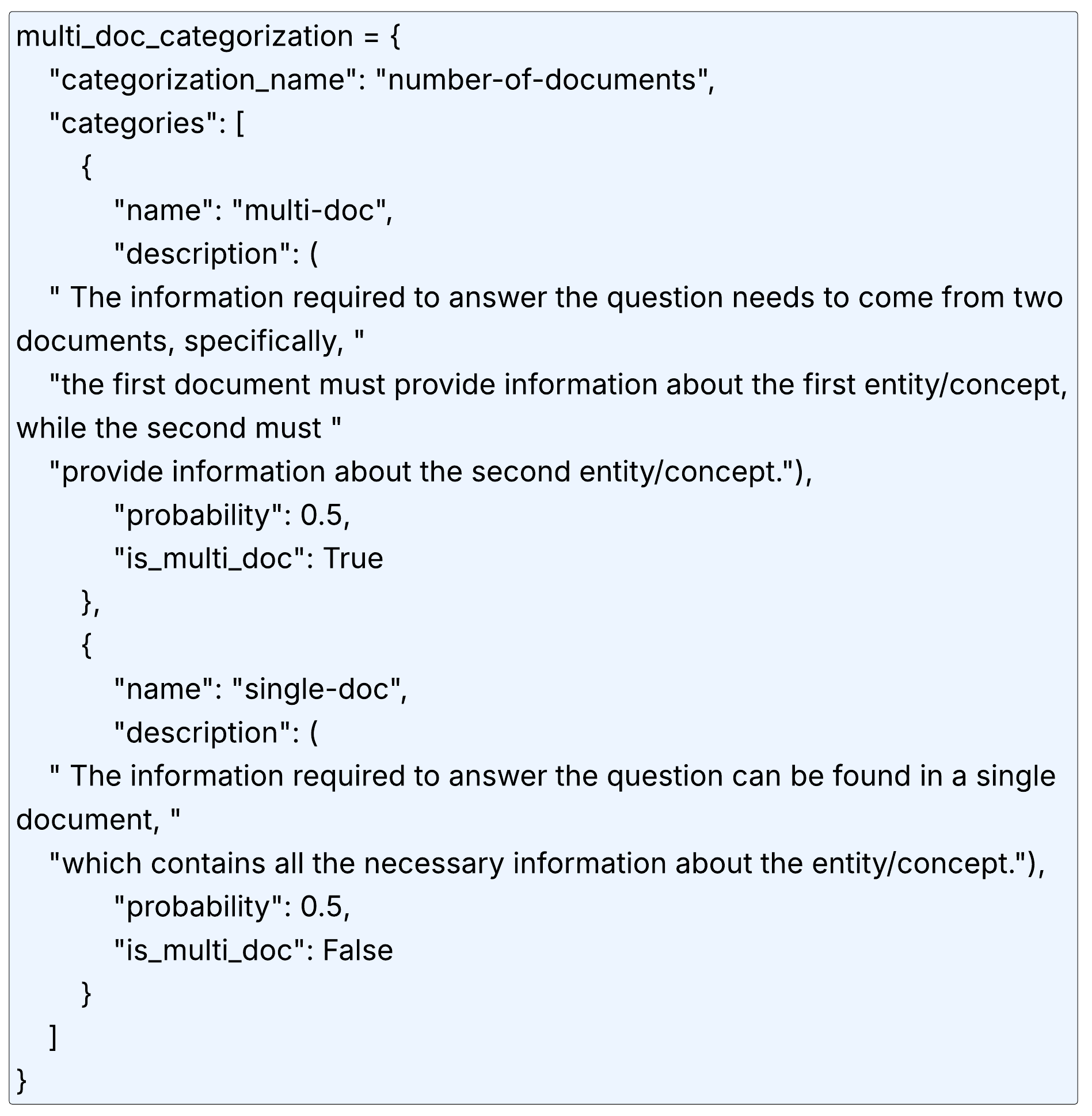}
    \caption{Document Granularity categorization schema and probabilities.}
    \label{fig:document-category}
\end{figure}

\begin{figure}
    \centering
    \includegraphics[width=1.0\linewidth]{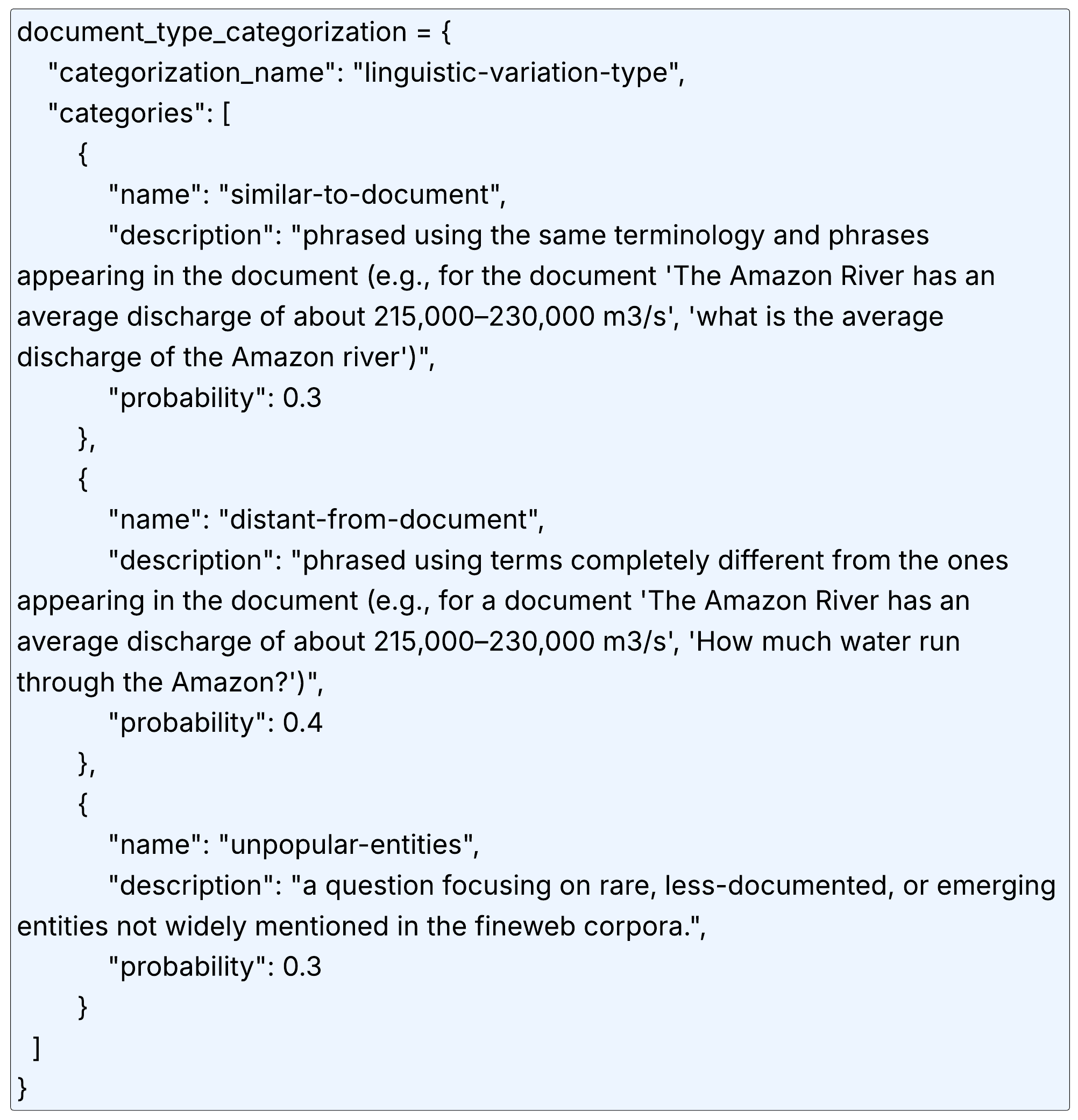}
    \caption{Lexical Similarity categorization schema and probabilities.}
    \label{fig:document-variation-category}
\end{figure}

This section provides verbatim definitions of all categorization schemas and their corresponding probability distributions used to control generation with \textsc{DataMorgana}. The following figures present the individual category schemata used to structure prompt-based question generation with \textsc{DataMorgana}:
\begin{itemize}[leftmargin=*]
    \item \textbf{User Expertise:} the definition of each subcategory and its probability of generating data with DataMorgana are shown in Figure~\ref{fig:user-category}.

    \item \textbf{Question Type:} the definition of each subcategory and its probability to generate data with DataMorgana are shown in Figure~\ref{fig:question-phrasing-category}.
    
    \item \textbf{Answer Type:} the definition of each subcategory and its probability of generating data with DataMorgana are shown in Figure~\ref{fig:answer-category}.
    
    \item \textbf{Question Intent:} the definition of each subcategory and its probability of generating data with DataMorgana are shown in Figure~\ref{fig:question-intent-category}.
    
    \item \textbf{Answer Intent:} the definition of each subcategory and its probability of generating data with DataMorgana are shown in Figure~\ref{fig:question-factuality-category}.
    
    \item \textbf{Premise Inclusion:} the definition of each subcategory and its probability of generating data with DataMorgana are shown in Figure~\ref{fig:question-premise-category}.
    
    \item \textbf{Lexical Similarity:} the definition of each subcategory and its probability of generating data with DataMorgana are shown in Figure~\ref{fig:document-variation-category}.
    
    \item \textbf{Aspect Granularity:} the definition of each subcategory and its probability of generating data with DataMorgana are shown in Figure~\ref{fig:question-aspect-category}.
    
    \item \textbf{Interaction Type:} the definition of each subcategory and its probability of generating data with DataMorgana are shown in Figure~\ref{fig:question-turn-category}.
    
    \item \textbf{Document Granularity:} the definition of each subcategory and its probability of generating data with DataMorgana are shown in Figure~\ref{fig:document-category}.
\end{itemize}

As noted before, we observed that jointly conditioning on all categories often led to incoherent or degraded outputs. To mitigate this, we designed the following five compatible category combinations for controlled generation:
\begin{enumerate}[leftmargin=*]
    \item[\textbf{(1)}]  User Expertise, Question Type, Answer Type, Document Granularity, Interaction Type, Aspect Granularity, Answer Intent, and Lexical Similarity.
    
    \item[\textbf{(2)}]  User Expertise, Question Type, Answer Type, Document Granularity, Interaction Type, Aspect Granularity, and Premise Inclusion.
    
    \item[\textbf{(3)}]  User Expertise, Question Type, Answer Type, Document Granularity, Interaction Type, Aspect Granularity, and Question Intent.
    
    \item[\textbf{(4)}]  User Expertise, Question Type, Answer Type, Document Granularity, Interaction Type, Aspect Granularity, Premise Inclusion, and Question Intent.
    
    \item[\textbf{(5)}]  User Expertise, Question Type, Answer Type, Document Granularity, Interaction Type, Aspect Granularity, Lexical Similarity, and Premise Inclusion.
\end{enumerate}

\section{Baselines Implementations Details}
\label{app:baselines}

\begin{figure}
    \centering
    \includegraphics[width=\linewidth]{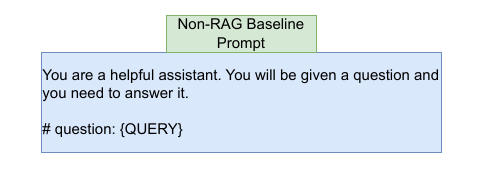}
    \vspace{-0.8cm}
    \caption{Prompt used with the Non-RAG baseline.}
    \label{fig:baseline-no-rag}
\end{figure}

\begin{figure}
    \centering
    \includegraphics[width=\linewidth]{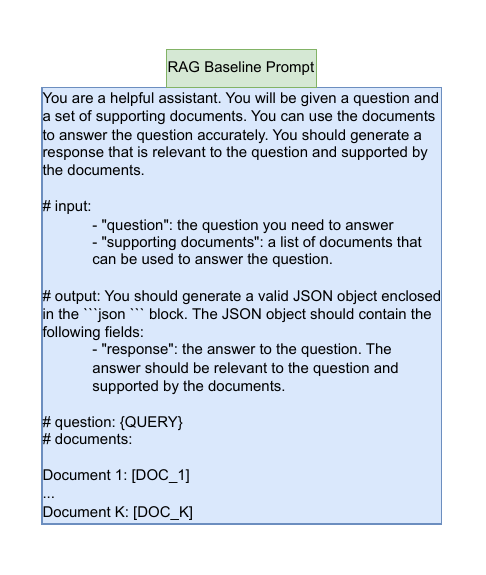}
    \vspace{-1.5cm}
    \caption{Prompt used with the RAG baseline that uses retrieve and read paradigm.}
    \label{fig:baseline-rag}
\end{figure}

As baselines in this paper, we utilize two distinct models:
\begin{itemize}[leftmargin=*]
    \item \textbf{No-Retrieval Baseline:} We use the same generator as in \ourmethod, namely the instruction-tuned Falcon model with 10B parameters\footnote{Available at: \url{https://huggingface.co/tiiuae/Falcon3-10B-Instruct}} \cite{Falcon3}, but without any retrieval. The model directly answers the question using the prompt shown in Figure~\ref{fig:baseline-no-rag}.

    \item \textbf{Vanilla RAG:} Again employing the instruction-tuned Falcon 10B model \cite{Falcon3}, this baseline incorporates retrieval. Documents are first retrieved using the same retrieval model and configuration as \ourmethod, with the question itself as the query. The retrieved documents are then passed along with the question to the LLM to generate a response using the prompt shown in Figure~\ref{fig:baseline-rag}.
\end{itemize}
where both baselines use nucleus sampling \cite{holtzman2020curiouscaseneuraltext} with a temperature of 0.1 for generation. For retrieval, we consistently retrieve two documents, similar to \ourmethod.

\section{Retriever Implementation Details}
\label{app:retrieval}

\subsubsection*{\textbf{Fineweb Corpus:}} To enable efficient indexing and document-level retrieval, we preprocessed the original FineWeb corpus\footnote{Available at: \url{https://huggingface.co/datasets/HuggingFaceFW/fineweb/viewer/sample-10BT}} by splitting long documents into smaller overlapping chunks. Each chunk was limited to 512 tokens with an 80-token overlap to preserve local coherence. This chunking strategy expanded the dataset from approximately 14 million original documents to nearly 29 million indexed units. For consistency during inference and indexing, we assigned sequential numerical IDs to these chunks (e.g., 1, 2, 3, ...), and stored a mapping between these IDs and the original FineWeb document identifiers for traceability.

\subsubsection*{\textbf{Retrieval Model:}} For our retrieval setup, we used a state-of-the-art sparse retrieval model, Lion, from Hugging Face \cite{zeng2025scalingsparsedenseretrieval}.\footnote{\url{https://huggingface.co/hzeng/Lion-SP-1B-llama3-marco-mntp}} This 1-billion-parameter model was used to index the documents and retrieve results, following the prior methodology from our lab \cite{zeng2025scalingsparsedenseretrieval}. This approach achieved state-of-the-art performance on the MS MARCO dataset and demonstrated comparable results on the FineWeb dataset.

\subsubsection*{\textbf{Indexing:}} During indexing, we encoded each document chunk into a sparse vector of dimension 128k (corresponding to the model's vocabulary size), retaining only non-zero token-weight pairs. These representations were stored in an inverted index on disk. The resulting index occupies approximately 571 GB of storage.

\subsubsection*{\textbf{Retrieval setup:}} At inference time, incoming queries were tokenized and encoded into sparse query vectors using the same model and projection head. Retrieval was performed via CPU-side sparse dot-product computation between the query vector and the inverted index, accelerated using Numba\footnote{Available at: \url{https://github.com/numba/numba}} for multi-threaded execution. We used a top-$k$ retrieval strategy with $k=2$ by default, and applied a threshold of 0.0 to filter low-scoring documents. The entire retrieval stack was exposed through a FastAPI\footnote{Available at: \url{https://github.com/fastapi}} service that supports both single and batched queries.

\section{\ourmethod's Implementations Details}

\subsection{Prompts and Algorithms}
\label{app:agents-prompts}

\begin{figure*}
    \centering
    \includegraphics[width=0.98\textwidth]{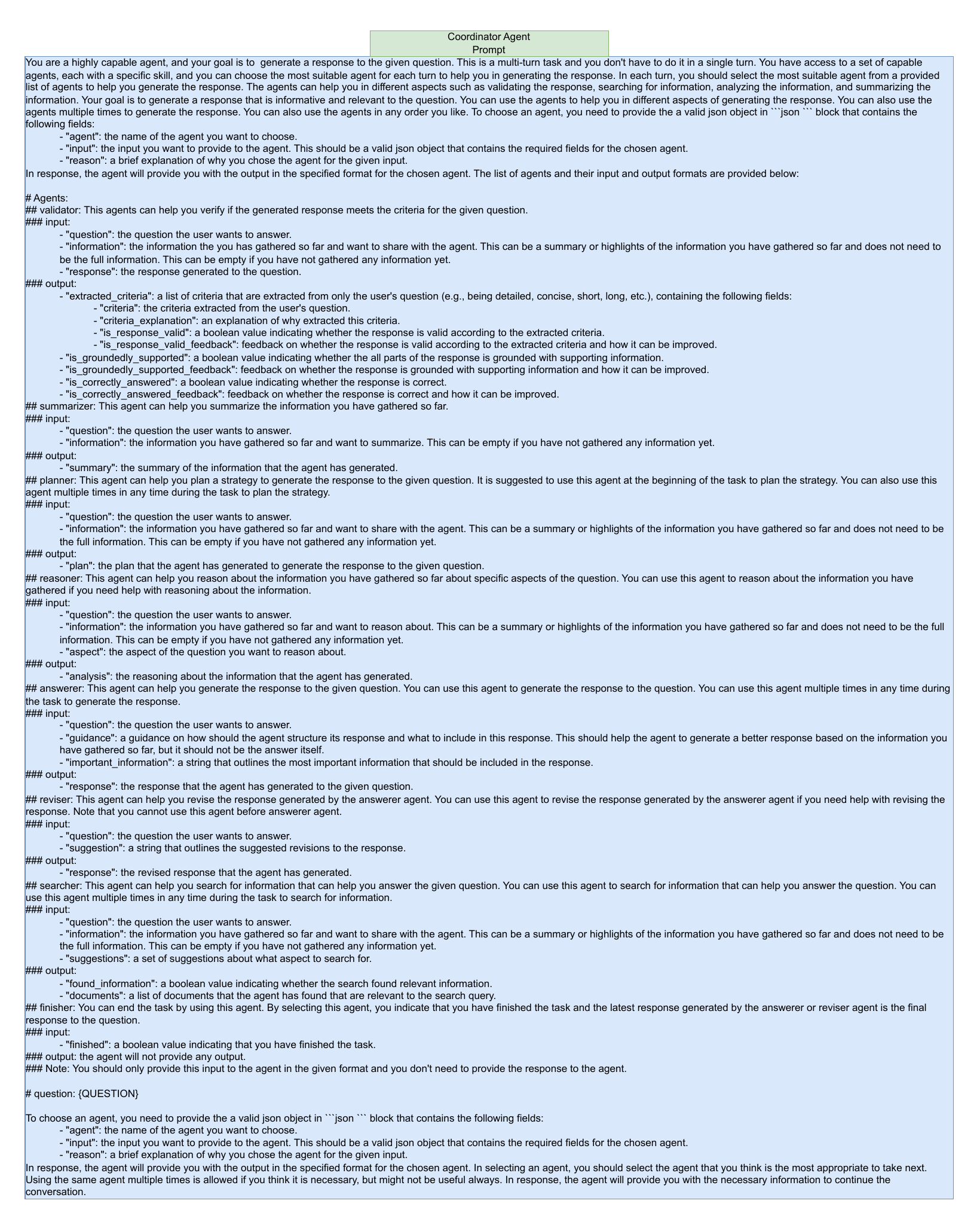}
    \vspace{-0.8cm}
    \caption{Prompt used for the Coordinator agent in the \ourmethod framework.}
    \label{fig:coordinator-agent}
\end{figure*}

\begin{figure*}
    \centering
    \includegraphics[width=\textwidth]{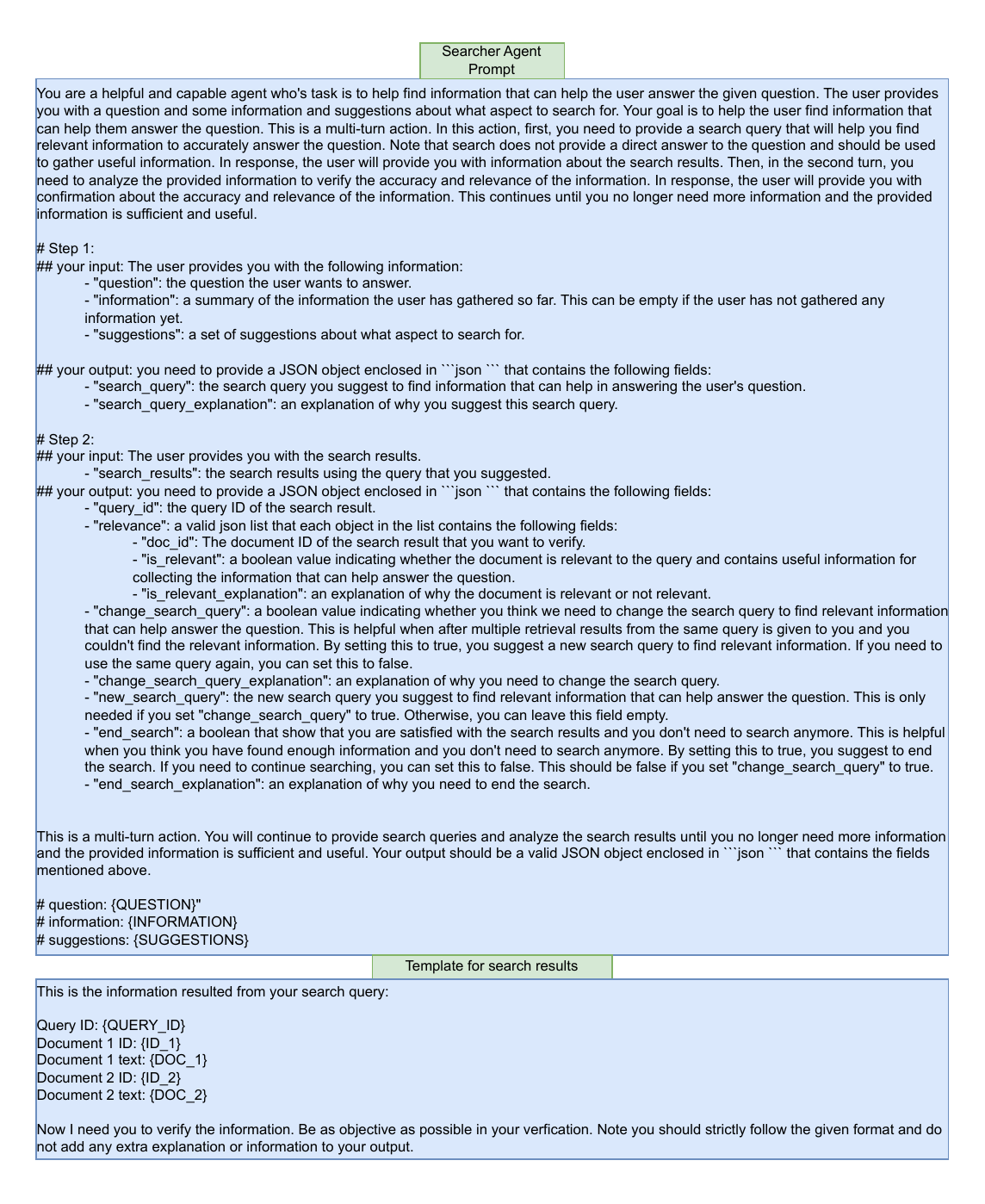}
    \vspace{-1.4cm}
    \caption{Prompt used for the Searcher agent in the \ourmethod framework.}
    \label{fig:searcher-agent}
\end{figure*}

\begin{figure*}
    \centering
    \includegraphics[width=0.85\textwidth]{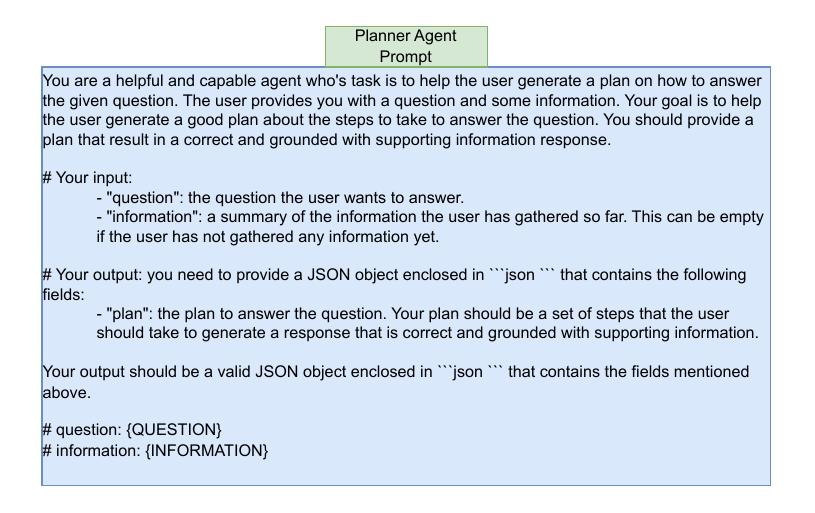}
    \vspace{-0.8cm}
    \caption{Prompt used for the Planner agent in the \ourmethod framework.}
    \label{fig:planner-agent}
\end{figure*}

\begin{figure*}
    \centering
    \includegraphics[width=0.85\textwidth]{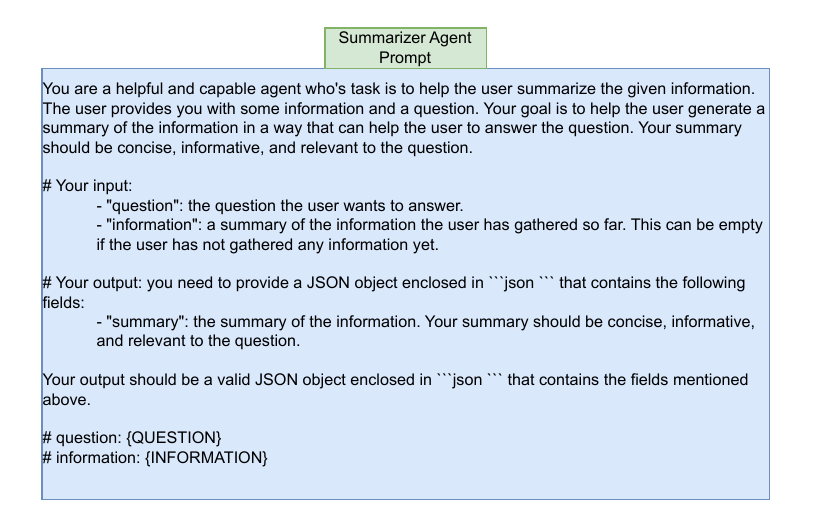}
    \vspace{-0.8cm}
    \caption{Prompt used for the Summarizer agent in the \ourmethod framework.}
    \label{fig:summarizer-agent}
\end{figure*}

\begin{figure*}
    \centering
    \includegraphics[width=0.85\textwidth]{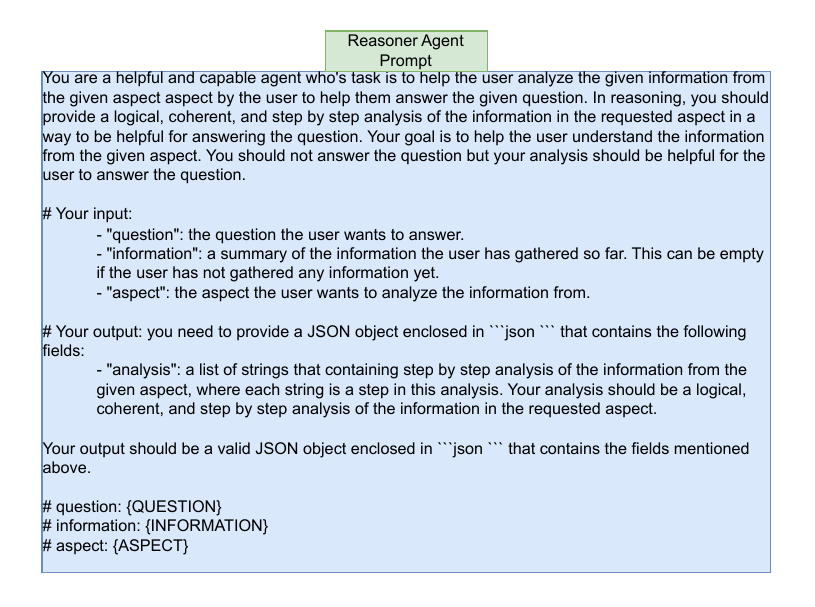}
    \vspace{-0.8cm}
    \caption{Prompt used for the Reasoner agent in the \ourmethod framework.}
    \label{fig:reasoner-agent}
\end{figure*}

\begin{figure*}
    \centering
    \includegraphics[width=0.85\textwidth]{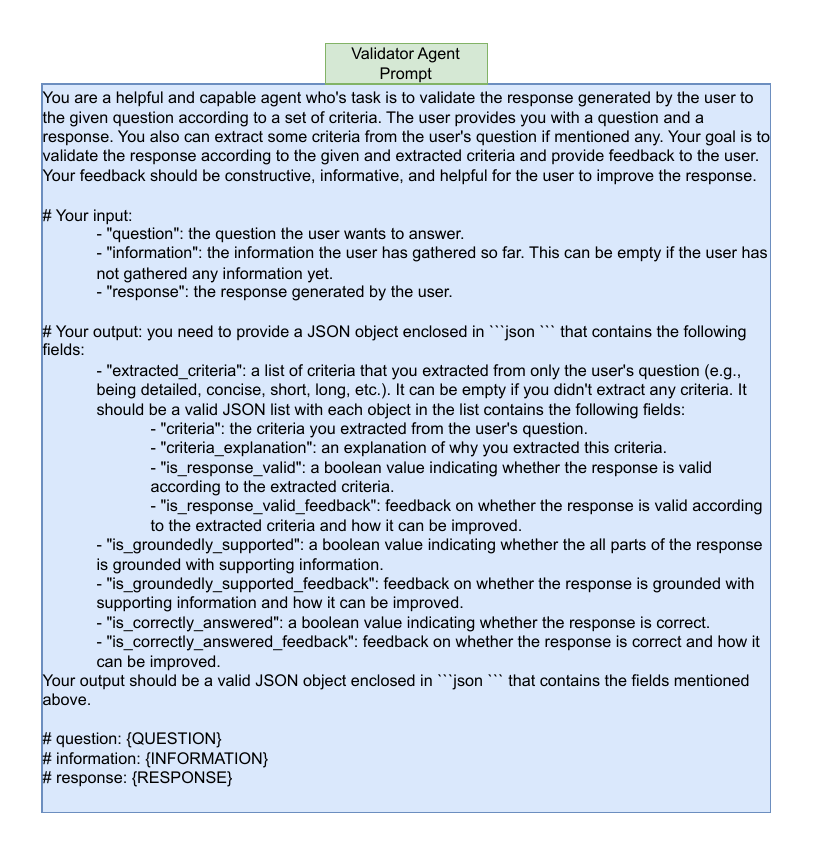}
    \vspace{-1cm}
    \caption{Prompt used for the Validator agent in the \ourmethod framework.}
    \label{fig:validator-agent}
\end{figure*}

\begin{figure*}
    \centering
    \includegraphics[width=0.85\textwidth]{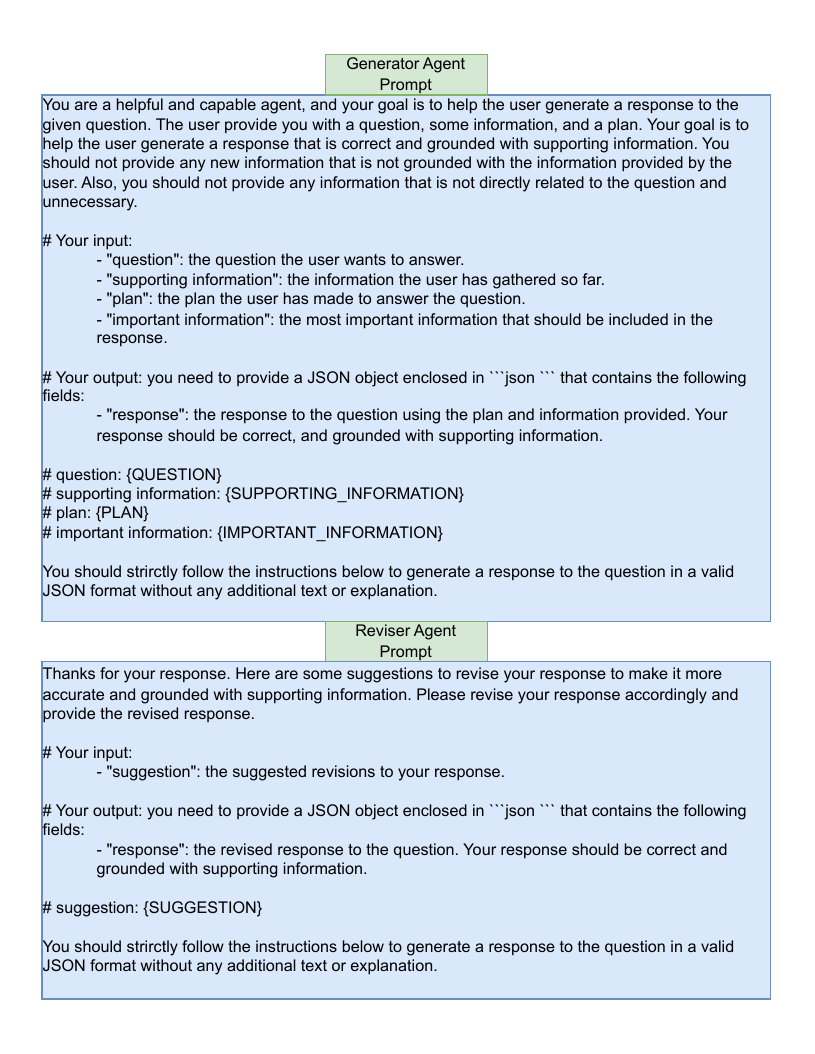}
    \vspace{-1cm}
    \caption{Prompt used for the Generator/Reviser agent in the \ourmethod framework.}
    \label{fig:generator-reviser-agent}
\end{figure*}

This section outlines the implementation details of the agents that collectively constitute \ourmethod. Each agent is instantiated with a specific role and is prompted accordingly using either the Qwen 2.5 model with 7B parameters\footnote{Available at: \url{https://huggingface.co/Qwen/Qwen2.5-7B-Instruct}} \cite{qwen2025qwen25technicalreport} or, in the case of response generation and revision, the Falcon 3 model with 10B parameters \cite{Falcon3}. The coordinator agent, whose prompt and logic are detailed in Figure~\ref{fig:coordinator-agent} and Algorithm~\ref{alg:coordinator}, manages the overall workflow by dispatching tasks to other agents based on context and agent specialization. It iteratively processes agent outputs, updating the conversation history and deciding when to terminate with a final answer. The planner (Figure~\ref{fig:planner-agent}) generates a high-level sequence of reasoning steps, while the searcher (Figure~\ref{fig:searcher-agent}, Algorithm~\ref{alg:searcher}) retrieves relevant documents using the Lion retrieval model \cite{zeng2025scalingsparsedenseretrieval} and dynamically adapts its search strategy. To support scalability and maintain coherence, the summarizer agent (Figure~\ref{fig:summarizer-agent}) compresses accumulated content, aiding the coordinator in maintaining context. When in-depth analytical reasoning is required, the reasoner (Figure~\ref{fig:reasoner-agent}) is invoked to process information on a specific aspect. The validator (Figure~\ref{fig:validator-agent}) ensures that the generated response meets all question-specific criteria by analyzing alignment between the question and the output. Finally, the generator/reviser agent (Figure~\ref{fig:generator-reviser-agent}), powered by Falcon 3 \cite{Falcon3}, produces and, if needed, refines the response to ensure completeness and coherence. Together, the agents operate in a tightly orchestrated loop, with their individual behaviors formalized in corresponding figures and algorithms.

\begin{algorithm*}
\caption{Implementation of the Coordinator agent in the \ourmethod framework.}\label{alg:coordinator}
\begin{algorithmic}[1]
\Require question $q$, agents list $A$, LLM $\pi$
\Ensure response to the question $r$, supporting documents $S$
\State $i = 0$ \Comment{Counter of number of steps/calls to agents}
\State ${finish} = False$ \Comment{Initializing variables for when to end process}
\State ${agents}\_{outputs} = []$ \Comment{Initializing variables for saving the state of agents}
\State $S = \{\}$ \Comment{Initializing variables for collecting supporting documents}
\State $r = ""$ \Comment{Initializing variables for final response}
\While{$i < \text{MAX\_STEPS}$ and ${finish} = False$} \Comment{As long as the agent wants to continue and has budget}
\State ${agent}\_{name}, {agent}\_{inputs} = \pi(q, {agents}\_{outputs}, S)$ \Comment{Selecting the next agent to be called and with what inputs}
\If{${agent}\_{name} = Finish$} \Comment{If coordinator chooses to finish response generation}
\State \textbf{Break} \Comment{End response generation}
\EndIf
\State $a = \text{select\_agent}(A, {agent}\_{name})$ \Comment{Select the agent from agents list}
\State ${selected}\_{agent}\_{output} = a({agent}\_{inputs})$ \Comment{Call the agent with the given input parameters generated by coordinator}
\State ${agents}\_{outputs} = {agents}\_{outputs} + [{selected}\_{agent}\_{output}]$ \Comment{Updating the state of the coordinator with the agent's output by appending}
\State $i = i + 1$ \Comment{Updating the step}
\If{${agent}\_{name} = {generator}$ or ${agent}\_{name} = {reviser}$} \Comment{If the selected agent generates a new response i.e., generator or reviser}
\State $r = {agents}\_{outputs}$ \Comment{Replace the response with the new response}
\EndIf
\If{${agent}\_{name} = {searcher}$} \Comment{If the selected agent collects supporting information i.e., searcher agent}
\State $S = S \cup {agents}\_{outputs}$ \Comment{Add the new supporting information to all supporting information}
\EndIf
\EndWhile
\State \Return $r$, $S$ \Comment{Returning the response and supporting documents}
\end{algorithmic}
\end{algorithm*}

\begin{algorithm*}
\caption{Implementation of the Searcher agent in the \ourmethod framework.}\label{alg:searcher}
\begin{algorithmic}[1]
\Require question $q$, context information $c$, suggested search aspects $a$, retrieval model $R$, LLM $\pi$
\Ensure relevant retrieved documents set $S$
\State $S = \{\}$ \Comment{Initializing the relevant documents set}
\State $i = 0$ \Comment{Counter of number of retrieval steps}
\State ${end}\_{search} = False$
\State ${query} = \pi(q, c, a)$
\While{$i < \text{MAX\_STEPS}$ and $end\_{search} = False$}
\State ${this}\_{step}\_{docs} =  R({query, 2)}$  \Comment{retrieve two docs for query. If this query is issued before, return next 2 docs for query.}
\For{$d \in {this}\_{step}\_{docs}$} \Comment{For each document in the retrieved document for the query}
    \If{$\pi({query, d}) = \text{relevant}$} \Comment{If document is relevant to the query}
    \State $S = S \cup \{d\}$ \Comment{Add the document to the relevant documents set}
    \EndIf
\EndFor
\State $i = i + 1$ \Comment{Updating the step}
\State ${change}\_{query} = \pi(query, {this}\_{step}\_{docs}, S)$ \Comment{Check if we need to change the query}
\State ${end}\_{search} =  \pi(query, {this}\_{step}\_{docs}, S)$ \Comment{Check if we need to end the search process}
\If{${change}\_{query} = True$}
\State ${query} = \pi(query, {this}\_{step}\_{docs}, S)$ \Comment{Updating the search query if it needs update}
\EndIf
\EndWhile
\State \Return $S$ \Comment{Returning the relevant retrieved documents}
\end{algorithmic}
\end{algorithm*}

\subsection{Training Setup}
\label{app:train-details}

To train the model, we first sample $T = 8$ diverse trajectories for each input in the training set. For all agents except the response generator/reviser, we use a temperature of 0.7 using nucleus sampling to encourage exploration and promote trajectory diversity. The response generator/reviser agent is sampled using nucleus sampling \cite{holtzman2020curiouscaseneuraltext} with a temperature of 0.1, as it is not trainable due to competition constraints. The lower temperature reduces randomness, leading to more deterministic outputs, which allows the trainable agents to better adapt to the fixed behavior of the generator/reviser.

To train the agents, we use a single LoRA \cite{hu2022lora} adaptor shared across all trainable agents, rather than a separate adaptor for each agent. This adaptor has a rank of $r = 16$ and is used to optimize all linear layers in the instruction-tuned Qwen 2.5 with 7B parameters LLM \cite{qwen2025qwen25technicalreport}. We use Unsloth\footnote{Available at: \url{https://github.com/unslothai/unsloth}} for efficient and accelerated training. The Adam optimizer \citep{adam} is employed with a learning rate of \(10^{-4}\). Gradient clipping with a maximum norm of 1 is applied. Training proceeds for up to 5,000 steps, including a 50-step warmup phase following a linear learning rate schedule. Models are evaluated every 500 steps on the test set, and the checkpoint with the best performance is selected. The combined maximum input and output length is set to 16,000 tokens. The batch size for all experiments is 128. All training is conducted on a single NVIDIA A100 GPU.

\subsection{Reward Models}
\label{app:train-reward}

\begin{figure*}
    \centering
    \includegraphics[width=0.85\textwidth]{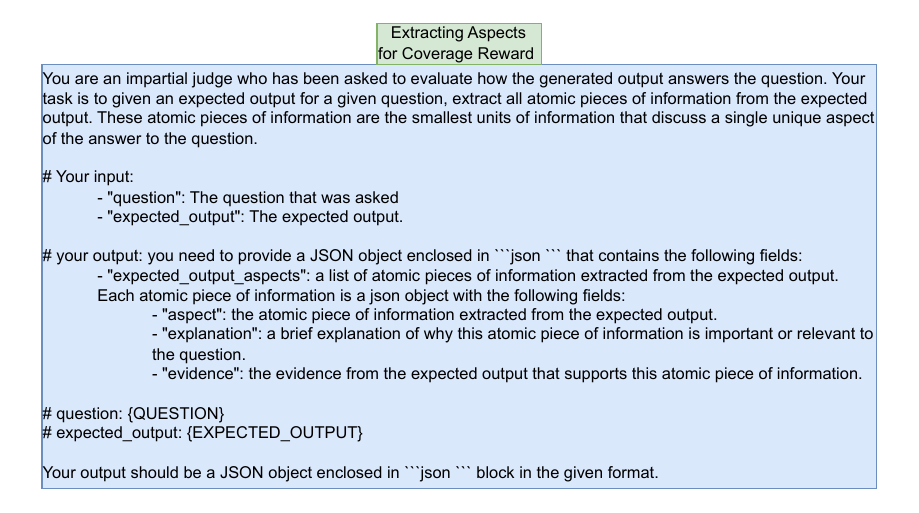}
    \vspace{-0.8cm}
    \caption{Prompt used for extracting atomic aspects from the expected output for recall-oriented nugget correctness reward model.}
    \label{fig:correctness-extract}
\end{figure*}

\begin{figure*}
    \centering
    \includegraphics[width=0.85\textwidth]{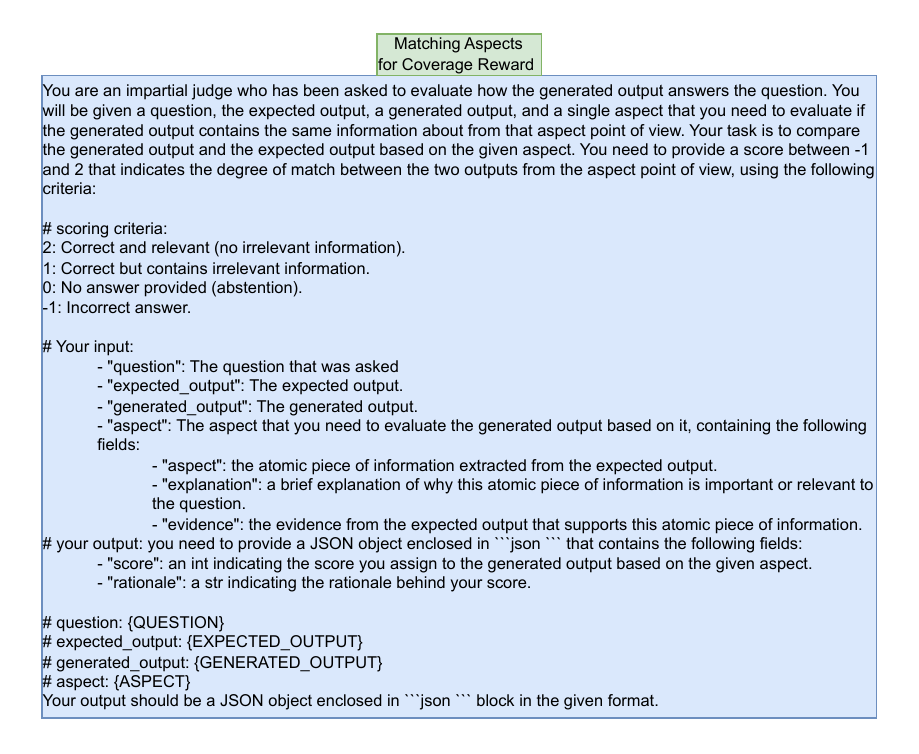}
    \vspace{-0.8cm}
    \caption{Prompt used for matching atomic aspects between the expected output and generated response for recall-oriented nugget correctness reward model.}
    \label{fig:correctness-match}
\end{figure*}

\begin{figure*}
    \centering
    \includegraphics[width=0.85\textwidth]{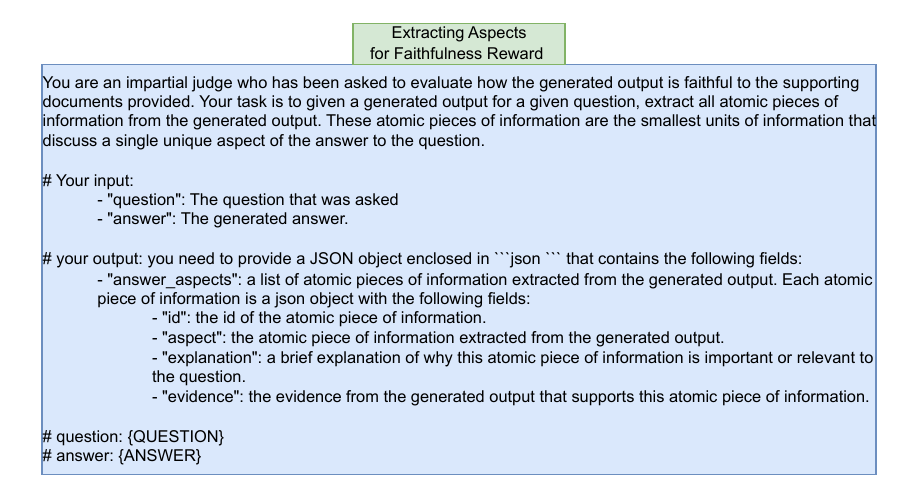}
    \vspace{-0.8cm}
    \caption{Prompt used for extracting atomic aspects from the generated output for faithfulness reward.}
    \label{fig:faithfulness-extract}
\end{figure*}

\begin{figure*}
    \centering
    \includegraphics[width=0.85\textwidth]{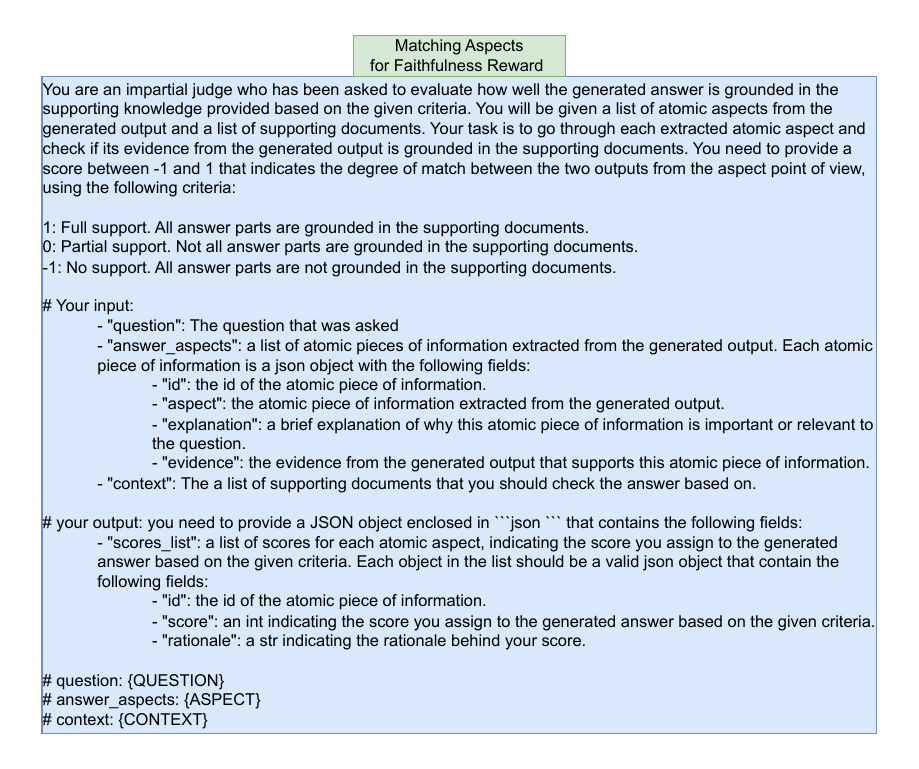}
    \vspace{-0.8cm}
    \caption{Prompt used for matching aspects between the generated output and retrieved documents for faithfulness reward.}
    \label{fig:faithfulness-match}
\end{figure*}

The competition does not provide an official scoring function, so we define and implement two complementary reward signals---correctness and faithfulness---based on the provided evaluation guidelines. correctness is assessed using a recall-oriented nugget-based reward following \citet{pradeep2025greatnuggetrecallautomating}, wherein atomic aspects are first extracted from the ground truth (Figure~\ref{fig:correctness-extract}) and then aligned with the generated response using a scoring prompt (Figure~\ref{fig:correctness-match}). These scores, normalized to [0, 1], are averaged to produce the final correctness reward, with the full implementation provided in Algorithm~\ref{alg:correctness}. Faithfulness follows the methodology of \citet{ragas}, involving the extraction of atomic claims (Figure~\ref{fig:faithfulness-extract}) and verifying their support against the retrieved documents (Figure~\ref{fig:faithfulness-match}), with final scores computed as the average normalized score (Algorithm~\ref{alg:faithfulness}). To reduce variance and ensure stable reward modeling and estimation, each reward model is executed five times with a temperature of 0.5 sampled using Nucleus Sampling \cite{holtzman2020curiouscaseneuraltext}, and the scores are averaged. Empirically, we observe that \ourmethod underperforms on the correctness dimension. To prioritize improvement in this area, we apply a weighting scheme in the final reward computation: a weight of 4 is assigned to correctness, while faithfulness receives a weight of 1. The final reward is then computed as the weighted average of the two normalized scores, effectively emphasizing correctness in the optimization process to improve it more. We use instruction-tuned Qwen 2.5 with 14 billion parameters\footnote{Available at: \url{https://huggingface.co/Qwen/Qwen2.5-14B-Instruct}} \cite{qwen2025qwen25technicalreport} as the backbone LLM for all the reward functions.

\subsection{Inference Setup}
\label{app:inference-setup}

For inference with LLMs, we utilize vLLM library\footnote{Available at: \url{https://github.com/vllm-project/vllm}} to serve the language models. A temperature of 0.1 with nucleus sampling is used for generation. The maximum combined input-output token length per agent is set to 32,000 tokens. The coordinator agent has a budget of up to 30 calls to other agents. The searcher agent can reuse the same query up to 5 times, with no additional budget constraints. Details of the retrieval model setup are provided in Appendix~\ref{app:retrieval}.

\section{Case Studies}
\label{app:case-study}

The proposed multi-agent system demonstrates notable and impressive behaviors, particularly in the decision-making processes of the coordinator and searcher agents, which are tasked with more complex responsibilities compared to others. To explore these behaviors in greater depth, we examine two representative case studies.

\subsubsection*{\textbf{Multi-aspect questions}}

In cases where the input question encompasses multiple dimensions and necessitates the collection of information across diverse aspects, the system must effectively retrieve and integrate information from these various dimensions to produce a coherent and high-quality response. One example of such a query is ``safety concerns in hydrogen steam reforming,'' which requires the system to identify and synthesize information across multiple safety-related aspects of the process. The full interaction between agents in response to this question is presented in Table~\ref{case-study-1}.

To address this question, the coordinator—serving as the entry point to the system---first invokes the planner agent to decompose the problem into a sequence of actionable steps. The planner breaks the task down into identifying the key components of hydrogen steam reforming, gathering information on each component, searching for associated safety hazards, and synthesizing the collected information into a coherent response. Subsequently, the coordinator directs the searcher agent, providing suggestions to guide the retrieval of information related to the specified aspects. The searcher then performs targeted queries and returns relevant information accordingly. After analyzing the safety aspects of the process using the reasoner agent, the coordinator instructs the searcher to gather information about real-world historical safety incidents related to hydrogen steam reforming. The searcher employs the retrieval model to obtain relevant documents and returns the collected data to the coordinator. Given the volume of retrieved information, the coordinator then invokes the summarizer agent to condense the content into a coherent form. Subsequently, the coordinator calls the generator agent, providing the aggregated evidence and instructing it to generate a comprehensive response. The generator is specifically guided to include a detailed overview of the key components involved in hydrogen steam reforming, the associated safety hazards for each component, documented historical incidents, and current safety guidelines and regulations. Once the response is generated, the coordinator consults the validator agent to verify whether the response meets the defined criteria. The validator emphasizes the importance of completeness and depth, particularly due to the safety-critical nature of the query. After confirming that all criteria are satisfied, the coordinator ends the process and returns the response.

\begin{algorithm*}
\caption{Implementation of the correctness reward model.}\label{alg:correctness}
\begin{algorithmic}[1]
\Require question $q$, generated response $r$, ground truth response $gt$, evaluator LLM $\pi$
\Ensure correctness score $s_{f}$
\State ${extracted}\_{aspects} = \pi(q, gt)$ \Comment{Extracting the atomic aspects from the ground truth output}
\State $S_{f} = 0$ 
\For{${aspect} \in {extracted}\_{aspects}$} \Comment{For each of the extracted aspects from the ground truth response}
\State $S_{aspect} = \frac{\pi(q, {aspect}, r, gt) + 1}{3}$ \Comment{Score the aspect using the LLM in range of -1 and 2, then normalize it to 0 and 1}
\State $S_{f} = S_{f} + \frac{S_{aspect}}{|{extracted}\_{aspects}|}$ \Comment{Adding the score of aspect to the final score}
\EndFor
\State \Return $s_f$ \Comment{Returning the correctness score for the generated response}
\end{algorithmic}
\end{algorithm*}

\begin{algorithm*}
\caption{Implementation of the Faithfulness reward model.}\label{alg:faithfulness}
\begin{algorithmic}[1]
\Require question $q$, generated response $r$, retrieved documents $C$, evaluator LLM $\pi$
\Ensure faithfulness score $s_{f}$
\State ${extracted}\_{aspects} = \pi(q, r)$ \Comment{Extracting the atomic aspects from the generated output}
\State $S_{f} = 0$ 
\For{${aspect} \in {extracted}\_{aspects}$} \Comment{For each of the extracted aspects from the generated response}
\State $S_{aspect} = \frac{\pi(q, {aspect}, r, C) + 1}{2}$ \Comment{Score the aspect using the LLM in range of -1 and 1, then normalize it to 0 and 1}
\State $S_{f} = S_{f} + \frac{S_{aspect}}{|{extracted}\_{aspects}|}$ \Comment{Adding the score of aspect to the final score}
\EndFor
\State \Return $s_f$ \Comment{Returning the faithfulness score for the generated response}
\end{algorithmic}
\end{algorithm*}

\subsubsection*{\textbf{Retrieving more information to make sure enough information is collected}}

In addition to cases that require integrating multiple aspects, some queries present challenges in retrieving relevant information even for a single aspect, often due to limitations in the retrieval model. In such situations, a robust RAG system must be capable of detecting retrieval failures and proactively reformulating search queries to obtain more relevant content. To demonstrate this capability, we examine the system’s response to the query: ``How did John Ball influence golf history at Hoylake, and what strategic challenges does the course present regarding out of bounds?'', as illustrated in Table~\ref{case-study-2}.

To address this question, the coordinator first invokes the planner agent to generate a structured plan. The resulting plan includes the following steps: identifying John Ball's role in golf history, retrieving key events or milestones in his career at Hoylake, and examining the evolution and impact of out-of-bounds (OB) rules on gameplay through historical records and expert analysis. Guided by this plan, the coordinator then tasks the searcher agent with retrieving information about John Ball’s influence on golf history at Hoylake, which is successfully completed. Subsequently, the reasoner agent is called to analyze John Ball’s contributions. During this reasoning process, it identifies the need to gather additional information regarding the strategic challenges of the Hoylake course. However, the initial retrieval attempt fails to yield relevant content. To address this, the searcher reformulates the query to focus on the “strategic challenges of Hoylake Royal Liverpool Golf Club,” which successfully retrieves pertinent information. The coordinator then invokes the generator agent to produce a comprehensive response using the collected evidence. As in the previous case study, this response is passed to the validator agent for evaluation. Once all criteria are satisfied, the coordinator concludes the process and returns the final response.

\input{case-study-1}
\input{case-study-2}

\end{document}

%% file: XX-notation.tex
\newcommand{\ourmethod}{{mRAG}\xspace}

%% file: case-study-1.tex
\clearpage
\onecolumn

\begin{longtable}{|p{2.8cm}|p{12.2cm}|}
\caption{First case study of \ourmethod on the question: safety concerns hydrogen steam reforming \label{case-study-1}} \\
\hline
\rowcolor{gray!20}
\textbf{Agent} & \textbf{Conversation} \\
\textbf{Coordinator} & 
\begin{tabular}[t]{@{}p{12cm}@{}}
\raggedright
\textbf{User:} \{"question": "safety concerns hydrogen steam reforming"\}
\rule{\linewidth}{0.4pt}
\textbf{LLM:} \{"agent": "planner",\\ "input": \{"question": "safety concerns hydrogen steam reforming", "information": ""\},\\ "reason": "I will use the planner agent to create a strategy for gathering and processing information related to safety concerns associated with hydrogen steam reforming." \}
\end{tabular} \\
\hline

\textbf{Planner} & 
\begin{tabular}[t]{@{}p{12cm}@{}}
\raggedright
\textbf{User (Coordinator):}\{
"question": "safety concerns hydrogen steam reforming","information":""\}
\rule{\linewidth}{0.4pt}
\textbf{LLM (Planner):} \{
"plan": ["Identify the key components of hydrogen steam reforming process", "Research potential safety hazards associated with each component", "Gather data on accidents or incidents related to hydrogen steam reforming from reliable sources such as industry reports, academic journals, and government databases", "Review existing safety guidelines and regulations for hydrogen steam reforming from organizations like OSHA, ISO, or relevant national agencies", "Analyze the risk factors and propose mitigation strategies for identified safety concerns", "Summarize findings and present a comprehensive overview of safety concerns and recommended measures"]
\}
\end{tabular} \\
\hline

\textbf{Coordinator} & 
\begin{tabular}[t]{@{}p{12cm}@{}}
\raggedright
\textbf{User(Planner):} "Output from Planner [Avoiding Repetition]"
\rule{\linewidth}{0.4pt}
\textbf{LLM:} \{"agent": "searcher", "input": \{    "question": "safety concerns hydrogen steam reforming",    "information": "",    "suggestions": ["key components of hydrogen steam reforming process","potential safety hazards associated with each component","accidents or incidents related to hydrogen steam reforming","existing safety guidelines and regulations for hydrogen steam reforming"  ]\},\\ "reason": "The searcher agent will help me find relevant information about the key components of hydrogen steam reforming, potential safety hazards, historical incidents, and current safety guidelines."
\}
\end{tabular} \\
\hline

\textbf{Searcher} & 
\begin{tabular}[t]{@{}p{12cm}@{}}
\raggedright
\textbf{User (Coordinator):} "Output from Coordinator [Avoiding Repetition]"
\rule{\linewidth}{0.4pt}
\textbf{LLM (Searcher):} \{"search\_query": ""key components of hydrogen steam reforming process",  \\
"search\_query\_explanation": "To understand the safety concerns of hydrogen steam reforming, it is important to first identify its key components as these components are likely to pose potential safety hazards."\} \\
\rule{\linewidth}{0.4pt}
\textbf{User (Retriever):} This is the information resulted from your search query:\\
    Query ID: 0 \\
    Documnet 1 ID: 552\\
    Document 1 text: Product gas from the steam reforming of the methane and naphtha contains equilibrium amounts of hydrogen,...\\
    Documnet 2 ID: 352\\
    Document 2 text: A final process stream is produced by steam reforming of residual...\\
    Now I need you to verify the information. Be as objective as possible in your verification. Note you should strictly follow the given format and do not add any extra explanation or information to your output. \\
\rule{\linewidth}{0.4pt}
\textbf{LLM (Searcher):} \{  "query\_id": 0,  
"relevance": [\{"doc\_id": 552,\\
"is\_relevant": true,\\
"is\_relevant\_explanation": "This document provides detailed information about the key components of the hydrogen steam reforming process, including the reformer, the temperature conditions, and the water-shift gas reactions. It is highly relevant to understanding the safety concerns of the process." \\
\{"doc\_id": 352,\\
"is\_relevant": false, \\
"is\_relevant\_explanation": "This document focuses on the advantages and economic benefits of the steam reforming process but does not provide specific details about the key components or potential safety hazards. Therefore, it is less relevant to the current query."\}],  "change\_search\_query": false,\\
"change\_search\_query\_explanation": "The initial search query yielded relevant documents that provide detailed information about the key components of the hydrogen steam reforming process. No immediate need to change the search query.", \\
"new\_search\_query": "", \\
"end\_search": false, \\
"end\_search\_explanation": "More information is needed to fully address the safety concerns of the hydrogen steam reforming process. Additional searches may be necessary to find details on potential safety hazards and existing safety guidelines.\}
\rule{\linewidth}{0.4pt}
\textbf{User (Retriever):} This is the information resulted from your search query:\\
Query ID: 0 \\
    Documnet 1 ID: 832\\
    Document 1 text: Steam Methane Reforming  one of the greatest challenges in today\u2019s industries...\\
    Documnet 2 ID: 915\\
    Document 2 text: 5,082,481, Membrane Separation Process for Cracked Gases, describes...\\
    Now I need you to verify the information. Be as objective as possible in your verification. Note you should strictly follow the given format and do not add any extra explanation or information to your output. \\
\rule{\linewidth}{0.4pt}
\textbf{LLM (Searcher):} \{  "query\_id": 0,  
"relevance": [\{"doc\_id": 832,\\
"is\_relevant": true,\\
"is\_relevant\_explanation": "This document provides detailed information about the steam methane reforming process, including the temperature and pressure conditions, the role of reformer tubes, and the importance of the process in hydrogen production. It is highly relevant to understanding the safety concerns of the process."\}\{"doc\_id": 915,\\
"is\_relevant": false,\\
"is\_relevant\_explanation": "This document discusses various membrane technologies for hydrogen recovery but does not provide specific details about the key components or potential safety hazards of the steam methane reforming process. Therefore, it is less relevant to the current query."\}],\\
\end{tabular}\\
\hline

\textbf{Searcher (Contd.)} & 
\begin{tabular}[t]{@{}p{12cm}@{}}
\raggedright
"change\_search\_query": false, \\
"change\_search\_query\_explanation": "The initial search query yielded relevant documents that provide detailed information about the steam methane reforming process. No immediate need to change the search query.", \\
"new\_search\_query": "", \\
"end\_search": false, \\
"end\_search\_explanation": "More information is needed to fully address the safety concerns of the steam methane reforming process. Additional searches may be necessary to find details on potential safety hazards and existing safety guidelines."\}
\rule{\linewidth}{0.4pt}
\textbf{User (Retriever):} This is the information resulted from your search query:\\
    Query ID: 0 \\
    Documnet 1 ID: 189\\
    Document 1 text: The reformer furnace design alternatives will be discussed below. Feed gas \u2014 usually a mixture of hydrogen, methane and other light hydrocarbons \u2014 is first compressed to about 300 psig....\\
    Documnet 2 ID: 254\\
    Document 2 text: BRIEF DESCRIPTION OF THE DRAWINGS These and other objects, features and advantages of the present invention will become better understood with regard to the following description, pending claims and accompanying drawings where:....\\
    Now I need you to verify the information. Be as objective as possible in your verification. Note you should strictly follow the given format and do not add any extra explanation or information to your output. 
\rule{\linewidth}{0.4pt}
\textbf{LLM (Searcher):} \{  "query\_id": 0,  
"relevance": [\{"doc\_id": 189,\\
"is\_relevant": true,\\
"is\_relevant\_explanation": "This document provides detailed information about the key components of the hydrogen steam reforming process, including the preheating, compression, and heating stages, as well as the role of reformer tubes and furnace designs. It is highly relevant to understanding the safety concerns of the process." \\
\{"doc\_id": 254,\\
"is\_relevant": false, \\
"is\_relevant\_explanation": "This document focuses on a novel pre-reforming and reforming process configuration involving membranes for hydrogen gas separation and recovery. While it mentions the steam methane reforming process, it does not provide specific details about the key components or potential safety hazards. Therefore, it is less relevant to the current query."\}],\\
"change\_search\_query": false,\\
"change\_search\_query\_explanation": "The initial search query yielded relevant documents that provide detailed information about the key components of the hydrogen steam reforming process. No immediate need to change the search query.", \\
"new\_search\_query": "", \\
"end\_search": false, \\
"end\_search\_explanation": "More information is needed to fully address the safety concerns of the hydrogen steam reforming process. Additional searches may be necessary to find details on potential safety hazards and existing safety guidelines.\}
\rule{\linewidth}{0.4pt}
\textbf{User (Retriever):} This is the information resulted from your search query:\\
    Query ID: 0 \\
    Documnet 1 ID: 985\\
    Document 1 text: Minimizing the hydrogen bled into the fuel gas can keep the hydrogen production levels manageable. However, the hydrogen plant feed usually includes some hydrogen that goes through for a \u201cfree ride,...\\
    Documnet 2 ID: 243\\
    Document 2 text: BRIEF DESCRIPTION OF THE DRAWINGS These and other objects, features and advantages of the present invention will become better understood with regard to the following description, pending claims and accompanying drawings where:....\\
\end{tabular}\\
\hline

\textbf{Searcher (Contd.)} & 
\begin{tabular}[t]{@{}p{12cm}@{}}
\raggedright
    Now I need you to verify the information. Be as objective as possible in your verification. Note you should strictly follow the given format and do not add any extra explanation or information to your output....
\rule{\linewidth}{0.4pt}
\textbf{LLM (Searcher):} \{  "query\_id": 0,  
"relevance": [\{"doc\_id": 189,\\
"is\_relevant": true,\\
"is\_relevant\_explanation": "This document provides detailed information about the key components of the hydrogen steam reforming process, including the preheating, compression, and heating stages, as well as the role of reformer tubes and furnace designs. It is highly relevant to understanding the safety concerns of the process." \\
\{"doc\_id": 254,\\
"is\_relevant": false, \\
"is\_relevant\_explanation": "This document focuses on a novel pre-reforming and reforming process configuration involving membranes for hydrogen gas separation and recovery. While it mentions the steam methane reforming process, it does not provide specific details about the key components or potential safety hazards. Therefore, it is less relevant to the current query."\}],\\
"change\_search\_query": false,\\
"change\_search\_query\_explanation": "The initial search query yielded relevant documents that provide detailed information about the key components of the hydrogen steam reforming process. No immediate need to change the search query.", \\
"new\_search\_query": "", \\
"end\_search": false, \\
"end\_search\_explanation": "More information is needed to fully address the safety concerns of the hydrogen steam reforming process. Additional searches may be necessary to find details on potential safety hazards and existing safety guidelines.\}
\rule{\linewidth}{0.4pt}
\textbf{User (Retriever):} This is the information resulted from your search query:\\
    Query ID: 0 \\
    Documnet 1 ID: 174\\
    Document 1 text: Gas enters the bottom of the absorber, where it contacts lean solution. The carbon dioxide is absorbed from the gas, leaving the rest of the contaminants and hydrogen relatively untouched.....\\
    Documnet 2 ID: 769\\
    Document 2 text: From our extensive range of gas production machinery, we manufacture, export and supply supreme quality Hydrogen Gas Generator.....\\
    Now I need you to verify the information. Be as objective as possible in your verification. Note you should strictly follow the given format and do not add any extra explanation or information to your output. 
\rule{\linewidth}{0.4pt}
\textbf{LLM (Searcher):} \{  "query\_id": 0,  
"relevance": [\{"doc\_id": 174,\\
"is\_relevant": false,\\
"is\_relevant\_explanation": "This document focuses on the PSA (Pressure Swing Adsorption) process for hydrogen purification and does not provide specific details about the key components or potential safety hazards of the steam methane reforming process. Therefore, it is less relevant to the current query." \\
\{"doc\_id": 769,\\
"is\_relevant": false, \\
"is\_relevant\_explanation": "This document describes a hydrogen gas generator and its features but does not provide specific details about the key components or potential safety hazards of the steam methane reforming process. Therefore, it is less relevant to the current query." "\}],\\
"change\_search\_query": true,\\
"change\_search\_query\_explanation": "The documents provided do not contain specific details about the key components or potential safety hazards of the steam methane reforming process. A new search query focusing on safety concerns and potential hazards is needed.",\\  "new\_search\_query": "potential safety hazards associated with steam methane reforming",\\ "end\_search": false,\\  "end\_search\_explanation": "More information is needed to fully address the safety concerns of the steam methane reforming process. A new search query focusing on potential safety hazards is necessary."
\}
\end{tabular} \\
\hline

\textbf{Searcher (Contd.)} & 
\begin{tabular}[t]{@{}p{12cm}@{}}
\raggedright
\textbf{User (Retriever):} This is the information resulted from your search query:\\
    Query ID: 0 \\
    Documnet 1 ID: 124\\
    Document 1 text: Gas enters the bottom of the absorber, where it contacts lean solution. The carbon dioxide is absorbed from the gas, leaving the rest of the contaminants and hydrogen relatively untouched.....\\
    Documnet 2 ID: 832\\
    Document 2 text: Steam Methane Reforming one of the greatest challenges in today\u2019s industries is an increased demand for hydrogen. Due to its ever-increasing industrial relevance, there is a need for efficient methods of hydrogen production, which has traditionally been done at a high temperature....\\
    Now I need you to verify the information. Be as objective as possible in your verification. Note you should strictly follow the given format and do not add any extra explanation or information to your output.
\rule{\linewidth}{0.4pt}
\textbf{LLM (Searcher):} \{  "query\_id": 1,  
"relevance": [\{"doc\_id": 124,\\
"is\_relevant": true,\\
"is\_relevant\_explanation": "This document provides detailed information about the potential safety hazards during the startup procedure of steam methane reformers, including the risk of overheating and melting of reformer tubes. It is highly relevant to understanding the safety concerns of the process." \\
\{"doc\_id": 832,\\
"is\_relevant": false, \\
"is\_relevant\_explanation": "This document focuses on the general description of the steam methane reforming process and its importance but does not provide specific details about potential safety hazards. Therefore, it is less relevant to the current query."\}],\\
"change\_search\_query": false,\\
"change\_search\_query\_explanation": "The initial search query yielded relevant documents that provide detailed information about the potential safety hazards during the steam methane reforming process. No immediate need to change the search query.",\\  "new\_search\_query": "",\\ "end\_search": true,\\  "end\_search\_explanation": "The provided information is sufficient to address the safety concerns of the steam methane reforming process. Further searches are not necessary." \}
\end{tabular}\\
\hline

\textbf{Coordinator} & 
\begin{tabular}[t]{@{}p{12cm}@{}}
\raggedright
\textbf{User (Searcher):} "only the relevant documents found by searcher"
\rule{\linewidth}{0.4pt}
\textbf{LLM :} \{  "agent": "reasoner",\\ 
"input": \{"question": "safety concerns hydrogen steam reforming", \\
"information": "Steam methane reforming is a chemical process by which methane is converted to hydrogen. It involves the reaction between natural gas, which contains methane and steam, at a temperature of 700-1000\u00b0C and under a pressure of 3-25 bar. The products of this chemical process are initially hydrogen and carbon monoxide. The carbon monoxide is further reacted with steam in a water-gas shift reaction to produce more hydrogen and carbon dioxide. Impurities are then removed from the system using pressure-swing adsorption. As a result, pure hydrogen is isolated and extracted. This process is mediated by reformer tubes \u2013 also known as catalyst tubes used in steam reformer furnaces known as steam methane reformers. Minimizing the hydrogen bled into the fuel gas can keep the hydrogen production levels manageable. However, the hydrogen plant feed usually includes some hydrogen that goes through for a free ride, except for the cost of heating it up to reformer temperature and ultimately cooling it back down to recovery level."\}, \\ "aspect": "Key components of hydrogen steam reforming process"\}, \\"reason": "I will use the reasoner agent to analyze the key components of the hydrogen steam reforming process and identify potential safety concerns associated with each component."\}
\end{tabular} \\
\hline

\textbf{Reasoner} & 
\begin{tabular}[t]{@{}p{12cm}@{}}
\raggedright
\textbf{User (Coordinator):} Output from Coordinator [Avoiding Repetition]
\rule{\linewidth}{0.4pt}
\textbf{LLM (Reasoner):} \{"analysis": "Hydrogen steam reforming involves multiple key components including the chemical reaction between methane and steam, the use of reformer tubes or catalyst tubes, and the subsequent purification steps.", "The primary chemical reaction occurs in the steam reformer furnace, where methane reacts with steam at high temperatures (700-1000\\u00b0C) and pressures (3-25 bar) to produce hydrogen and carbon monoxide.", "Reformer tubes or catalyst tubes play a crucial role in facilitating the chemical reaction. These tubes house the catalyst necessary for the steam reforming process, enhancing the rate of hydrogen production.", "After the initial reaction, a water-gas shift reaction takes place to further convert carbon monoxide into additional hydrogen and carbon dioxide, improving the purity of the hydrogen output.", "Purification steps such as pressure-swing adsorption are employed to remove impurities from the hydrogen stream, ensuring the final product is of high quality and suitable for various applications.", "Minimizing hydrogen bleed during the process helps maintain efficiency, but some hydrogen inevitably remains in the feed stream, requiring energy to heat and cool."\}
\end{tabular} \\
\hline

\textbf{Coordinator} & 
\begin{tabular}[t]{@{}p{12cm}@{}}
\raggedright
\textbf{User (Reasoner):} "Output from Reasoner [Avoiding Repetition]" 
\rule{\linewidth}{0.4pt}
\textbf{LLM :} \{  "agent": "reasoner",\\ 
"input": \{"question": "safety concerns hydrogen steam reforming", \\
"information": "Hydrogen steam reforming involves multiple key components including the chemical reaction between methane and steam, the use of reformer tubes or catalyst tubes, and the subsequent purification steps. The primary chemical reaction occurs in the steam reformer furnace, where methane reacts with steam at high temperatures (700-1000\u00b0C) and pressures (3-25 bar) to produce hydrogen and carbon monoxide. Reformer tubes or catalyst tubes play a crucial role in facilitating the chemical reaction. After the initial reaction, a water-gas shift reaction takes place to further convert carbon monoxide into additional hydrogen and carbon dioxide, improving the purity of the hydrogen output. Purification steps such as pressure-swing adsorption are employed to remove impurities from the hydrogen stream, ensuring the final product is of high quality and suitable for various applications. Minimizing hydrogen bleed during the process helps maintain efficiency, but some hydrogen inevitably remains in the feed stream, requiring energy to heat and cool."\},\\ "aspect": "Potential safety hazards associated with each component"\},\\ "reason": "I will use the reasoner agent to identify potential safety hazards associated with each key component of the hydrogen steam reforming process."\}
\end{tabular} \\
\hline

\textbf{Reasoner} & 
\begin{tabular}[t]{@{}p{12cm}@{}}
\raggedright
\textbf{User (Coordinator):} Output from Coordinator [Avoiding Repetition]
\rule{\linewidth}{0.4pt}
\textbf{LLM (Reasoner):} \{"analysis": "The steam reformer furnace is a critical component as it operates at extremely high temperatures (700-1000\\u00b0C). High temperatures pose potential risks such as thermal stress, cracking, and potential leaks of flammable gases like methane, leading to fire or explosion hazards.", "Reformer tubes or catalyst tubes are another key component. These tubes contain reactive materials that can degrade over time due to exposure to high temperatures and corrosive environments, potentially leading to tube failure and gas leaks. Additionally, the high-pressure environment (3-25 bar) increases the severity of any leaks, posing significant safety risks.", "The water-gas shift reactor, although not explicitly mentioned in the provided information, is an important step in the process. It involves the reaction of carbon monoxide with steam at around 300-400\\u00b0C to produce more hydrogen. High temperatures and the presence of reactive gases increase the risk of equipment failure and gas leaks, which could lead to inhalation of toxic gases or fires.", "Pressure-swing adsorption (PSA) is a purification step that uses adsorbents to remove impurities from the hydrogen stream. During the adsorption and desorption cycles, there is a risk of overpressure or vacuum conditions, which can damage the PSA equipment and lead to gas leaks. Furthermore, the handling of adsorbents can pose health risks if inhaled or contacted directly.", "Overall, the high temperatures, pressures, and flammable gases involved in the hydrogen steam reforming process mean that all components must be carefully designed, maintained, and monitored to ensure safety. Any failure in these systems can lead to serious safety hazards."\}
\end{tabular} \\
\hline

\textbf{Coordinator} & 
\begin{tabular}[t]{@{}p{12cm}@{}}
\raggedright
\textbf{User (Reasoner):} Output from Reasoner [Avoiding Repetition]
\rule{\linewidth}{0.4pt}
\textbf{LLM:} \{"agent": "searcher",\\  
"input": \{"question": "safety concerns hydrogen steam reforming",\\
"information": "The steam reformer furnace is a critical component as it operates at extremely high temperatures (700-1000\u00b0C). Reformer tubes or catalyst tubes are another key component. The water-gas shift reactor involves high temperatures and the presence of reactive gases. Pressure-swing adsorption (PSA) equipment handles overpressure and vacuum conditions.",\\"suggestions": ["Historical incidents related to hydrogen steam reforming","Existing safety guidelines and regulations for hydrogen steam reforming", "Mitigation strategies for identified safety concerns" ],\\
"reason": "The searcher agent will help me find historical incidents, existing safety guidelines, and mitigation strategies related to hydrogen steam reforming to provide a comprehensive overview of safety concerns and recommended measures."\}"
\end{tabular} \\
\hline

\textbf{Searcher} & 
\begin{tabular}[t]{@{}p{12cm}@{}}
\raggedright
\textbf{LLM (Searcher):} \{"search\_query": "historical incidents related to hydrogen steam reforming",  \\
"search\_query\_explanation": "Historical incidents can provide valuable insights into the types of safety issues that have occurred in the past, which can help identify current safety concerns and potential mitigation strategies."\}
\rule{\linewidth}{0.4pt}
\textbf{User (Retriever):} This is the information resulted from your search query:\\
    Query ID: 2 \\
    Documnet 1 ID: 785\\
    Document 1 text: Fuel refineries use a series of chemical reactions to arrive at the final product. Hydrogen fluoride is used in the oil refining process and causes a reaction, called alkylation, to make fuel from unrefined oil....\\
    Documnet 2 ID: 200\\
    Document 2 text:  Fig. 4: Mass fractions of the species (ethanol, CO2 and H2O) along the reformer axis.The mass fractions of the ethanol and steam decay along the reformer axis. The ethanol conversion is 80.3\%.....\\
    Now I need you to verify the information. Be as objective as possible in your verification. Note you should strictly follow the given format and do not add any extra explanation or information to your output.
\rule{\linewidth}{0.4pt}
\textbf{LLM (Searcher):} \{  "query\_id": 2,  
"relevance": [\{"doc\_id": 785,\\
"is\_relevant": true,\\
"is\_relevant\_explanation": "This document provides a historical incident related to a refinery accident involving hydrofluoric acid, which can be linked to safety concerns in steam reforming processes. It highlights the potential for catastrophic impacts and the importance of safety protocols."\\
\{"doc\_id": 200,\\
"is\_relevant": false, \\
"is\_relevant\_explanation": ""This document discusses the mass fractions of species in a steam reformer and safety concerns related to hydrogen integrity in steel tubes. While it provides technical details, it does not directly address historical incidents or safety guidelines specific to steam reforming. Therefore, it is less relevant to the current query."\}],\\
"change\_search\_query": false,\\
"change\_search\_query\_explanation": "The initial search query yielded relevant documents that provide a historical incident related to a refinery accident involving hydrofluoric acid, which can be linked to safety concerns in steam reforming processes. No immediate need to change the search query.",\\  "new\_search\_query": "",\\ "end\_search": true,\\  "end\_search\_explanation": "The provided information is sufficient to address the safety concerns of the steam methane reforming process. Further searches are not necessary." \}
\end{tabular} \\
\hline

\textbf{Coordinator} & 
\begin{tabular}[t]{@{}p{12cm}@{}}
\raggedright
\textbf{User (Searcher):} "only the relevant documents found by searcher"
\rule{\linewidth}{0.4pt}
\textbf{LLM (Coordinator):} \{"agent": "summarizer",\\
"input": \{"question": "safety concerns hydrogen steam reforming",\\
"information": ""The steam reformer furnace is a critical component as it operates at extremely high temperatures (700-1000\u00b0C). Reformer tubes or catalyst tubes are another key component. The water-gas shift reactor involves high temperatures and the presence of reactive gases. Pressure-swing adsorption (PSA) equipment handles overpressure and vacuum conditions. Historical incidents related to hydrogen steam reforming include an explosion at an ExxonMobil refinery in Torrance, California, where an explosion occurred due to a leak of hydrocarbons into an electrostatic precipitator (ESP) device, causing sparks and igniting the hydrocarbons. This resulted in a dust cloud extending for miles and projected large debris around the site. The incident highlighted the importance of proper evaluation of process hazards, implementation of safeguards, and adherence to safety precautions."\}\\
"reason": "I will use the summarizer agent to create a concise summary of the key points regarding the safety concerns associated with hydrogen steam reforming based on the gathered information."\}
\end{tabular} \\
\hline

\textbf{Summarizer} & 
\begin{tabular}[t]{@{}p{12cm}@{}}
\raggedright
\textbf{User (Coordinator):} Output from Coordinator and relevant retrieved documents [Avoiding Repetition]
\rule{\linewidth}{0.4pt}
\textbf{LLM (Summarizer):} \{"summary": "Hydrogen steam reforming involves critical components like high-temperature furnaces (700-1000\\u00b0C), reformer/tubes, and water-gas shift reactors. PSA equipment manages overpressure/vacuum conditions. Historical incidents, such as an explosion at ExxonMobil\\u2019s Torrance refinery, underscore the need for thorough hazard assessments, robust safety measures, and strict adherence to safety protocols."\}
\end{tabular} \\
\hline

\textbf{Coordinator} & 
\begin{tabular}[t]{@{}p{12cm}@{}}
\raggedright
\textbf{User (Answerer):} Output from Summarizer [Avoiding Repetition]
\rule{\linewidth}{0.4pt}
\textbf{LLM:}\{"agent": "Answerer",  \\
"input": \{"question": "safety concerns hydrogen steam reforming",\\  
"guidance": "Provide a detailed overview of the key components of hydrogen steam reforming, potential safety hazards associated with each component, historical incidents, and existing safety guidelines and regulations.","important\_information": "High-temperature furnaces (700-1000\u00b0C), reformer/tubes, water-gas shift reactors, PSA equipment, historical incidents, and safety guidelines."\}\\ 
"reason": "I will use the answerer agent to generate a comprehensive response to the question about safety concerns in hydrogen steam reforming, incorporating the key components, potential hazards, historical incidents, and safety guidelines."\},
\end{tabular} \\
\hline

\textbf{Answerer} & 
\begin{tabular}[t]{@{}p{12cm}@{}}
\raggedright
\textbf{User (Coordinator):} Output from Coordinator [Avoiding Repetition]
\rule{\linewidth}{0.4pt}
\textbf{LLM:} \{"response": "Hydrogen steam reforming is a critical process in the production of hydrogen, involving several key components such as high-temperature furnaces, reformer/tubes, water-gas shift reactors, and PSA equipment. Each component presents specific safety concerns. High-temperature furnaces, operating at 700-1000\\u00b0C, pose risks of thermal burns and potential tube failures due to excessive temperatures. Reformer/tubes are susceptible to coking and require careful monitoring to prevent catalyst degradation. Water-gas shift reactors, where the exothermic reaction occurs, need proper cooling systems to manage temperature increases. PSA equipment, used for hydrogen purification, must be handled with care to avoid pressure-related accidents. Historical incidents, such as the ExxonMobil refinery explosion involving hydrofluoric acid, highlight the importance of stringent safety measures. Safety guidelines emphasize the need for proper monitoring, regular maintenance, and adherence to operating procedures to mitigate risks. Regulations require thorough hazard assessments and the implementation of safeguards to prevent accidents. Understanding these components and their associated safety concerns is crucial for the safe operation of hydrogen steam reforming facilities." \}
\end{tabular} \\
\hline

\textbf{Coordinator} & 
\begin{tabular}[t]{@{}p{12cm}@{}}
\raggedright
\textbf{User (Answerer):} Output from Answerer [Avoiding Repetition] \rule{\linewidth}{0.4pt}
\textbf{LLM:}  \{"agent": "validator", "input": \{ "question":"safety concerns hydrogen steam reforming"\\ "information": "Hydrogen steam reforming is a critical process in the production of hydrogen, involving several key components such as high-temperature furnaces, reformer/tubes, water-gas shift reactors, and PSA equipment. Each component presents specific safety concerns. High-temperature furnaces, operating at 700-1000\u00b0C, pose risks of thermal burns and potential tube failures due to excessive temperatures. Reformer/tubes are susceptible to coking and require careful monitoring to prevent catalyst degradation. Water-gas shift reactors, where the exothermic reaction occurs, need proper cooling systems to manage temperature increases. PSA equipment, used for hydrogen purification, must be handled with care to avoid pressure-related accidents. Historical incidents, such as the ExxonMobil refinery explosion involving hydrofluoric acid, highlight the importance of stringent safety measures. Safety guidelines emphasize the need for proper monitoring, regular maintenance, and adherence to operating procedures to mitigate risks. Regulations require thorough hazard assessments and the implementation of safeguards to prevent accidents. Understanding these components and their associated safety concerns is crucial for the safe operation of hydrogen steam reforming facilities.",\\  "response": "Hydrogen steam reforming is a critical process in the production of hydrogen, involving several key components such as high-temperature furnaces, reformer/tubes, water-gas shift reactors, and PSA equipment. Each component presents specific safety concerns. High-temperature furnaces, operating at 700-1000\u00b0C, pose risks of thermal burns and potential tube failures due to excessive temperatures. Reformer/tubes are susceptible to coking and require careful monitoring to prevent catalyst degradation. Water-gas shift reactors, where the exothermic reaction occurs, need proper cooling systems to manage temperature increases. PSA equipment, used for hydrogen purification, must be handled with care to avoid pressure-related accidents. Historical incidents, such as the ExxonMobil refinery explosion involving hydrofluoric acid, highlight the importance of stringent safety measures. Safety guidelines emphasize the need for proper monitoring, regular maintenance, and adherence to operating procedures to mitigate risks. Regulations require thorough hazard assessments and the implementation of safeguards to prevent accidents. Understanding these components and their associated safety concerns is crucial for the safe operation of hydrogen steam reforming facilities."\},\\ "reason": "I will use the validator agent to check if the generated response meets the criteria for the given question."\}
\end{tabular} \\
\hline

\textbf{Validator} & 
\begin{tabular}[t]{@{}p{12cm}@{}}
\raggedright
\textbf{User (Coordinator):} Output from Coordinator [Avoiding Repetition]
\rule{\linewidth}{0.4pt}
\textbf{LLM:} \{"extracted\_criteria": [\{"criteria": "detailed", "criteria\_explanation": "The question asks about safety concerns related to hydrogen steam reforming, which requires a detailed explanation.", "is\_response\_valid\": true, "is\_response\_valid\_feedback": "The response is detailed and covers various aspects of safety concerns in hydrogen steam reforming."\}], \\
"is\_groundedly\_supported": true, "is\_groundedly\_supported\_feedback": "The response is well-grounded with supporting information, including specific components, potential risks, and historical incidents.", "is\_correctly\_answered": true, "is\_correctly\_answered\_feedback": "The response correctly addresses the safety concerns of hydrogen steam reforming and provides relevant details."\}"
\end{tabular}\\
\hline

\textbf{Coordinator} & 
\begin{tabular}[t]{@{}p{12cm}@{}}
\raggedright
\textbf{User (Validator):} Output from Validator [Avoiding Repetition]
\rule{\linewidth}{0.4pt}
\textbf{LLM:} \{"agent":"finisher", "input":{"finished":true}\}
\end{tabular}\\
\hline

\end{longtable}

\twocolumn

%% file: case-study-2.tex
\clearpage
\onecolumn

\begin{longtable}{|p{2.8cm}|p{12.2cm}|}
\caption{Second case study of \ourmethod on the question: How did John Ball influence golf history at Hoylake, and what strategic challenges does the course present regarding out of bounds? \label{case-study-2}} \\
\hline
\rowcolor{gray!20}
\textbf{Agent} & \textbf{Conversation} \\
\textbf{Coordinator} & 
\begin{tabular}[t]{@{}p{12cm}@{}}
\raggedright
\textbf{User:} \{"question": "How did John Ball influence golf history at Hoylake, and what strategic challenges does the course present regarding out of bounds?"\}
\rule{\linewidth}{0.4pt}
\textbf{LLM:} \{"agent": "planner",\\ "input": \{"question": "How did John Ball influence golf history at Hoylake, and what strategic challenges does the course present regarding out of bounds?", "information": ""\},\\ "reason": "I need to create a plan to gather and organize the necessary information to answer the question." \}
\end{tabular} \\
\hline

\textbf{Planner} & 
\begin{tabular}[t]{@{}p{12cm}@{}}
\raggedright
\textbf{User (Coordinator):}\{
"question": "How did John Ball influence golf history at Hoylake, and what strategic challenges does the course present regarding out of bounds?" \\
"information": "" \}
\rule{\linewidth}{0.4pt}
\textbf{LLM (Planner):} \{
"plan": ["Research John Ball's role in golf history, focusing on his connection to Hoylake Golf Club.", "Identify key events or moments in John Ball's career at Hoylake that significantly impacted golf history.", "Analyze the current layout and characteristics of the Hoylake course, particularly its out of bounds (OB) rules and challenges it presents to players.", "Consult historical records and expert opinions to understand how these OB rules have evolved over time and their impact on gameplay.", "Compile findings into a coherent narrative that highlights John Ball's influence and the strategic challenges posed by the course."]
\}
\end{tabular} \\
\hline

\textbf{Coordinator} & 
\begin{tabular}[t]{@{}p{12cm}@{}}
\raggedright
\textbf{User(Planner):} "Output from Planner [Avoiding Repetition]"
\rule{\linewidth}{0.4pt}
\textbf{LLM:} \{"agent": "searcher",\\  "input": {    "question": "How did John Ball influence golf history at Hoylake, and what strategic challenges does the course present regarding out of bounds?", "information": "", "suggestions": ["John Ball", "Hoylake Golf Club", "history of golf at Hoylake", "out of bounds rules in golf"]},\\ "reason": "The next step is to find relevant information about John Ball's connection to Hoylake and the characteristics of the Hoylake course."\}
\end{tabular} \\
\hline

\textbf{Searcher} & 
\begin{tabular}[t]{@{}p{12cm}@{}}
\raggedright
\textbf{User (Coordinator):} "Output from Coordinator [Avoiding Repetition]"
\rule{\linewidth}{0.4pt}
\textbf{LLM (Searcher):} \{"search\_query": "John Ball influence on Hoylake golf history",  \\
"search\_query\_explanation": "This search query aims to find information specifically related to how John Ball influenced golf history at Hoylake, which directly addresses part of the question."\} \\
\rule{\linewidth}{0.4pt}
\textbf{User (Retriever):} This is the information resulted from your search query:\\
    Query ID: 0 \\
    Documnet 1 ID: 224\\
    Document 1 text: United Kingdom: Grant Books, 2001. 81p. half leather, slipcase. Limited edition of 95 subscribers copies. Hoylake, otherwise known as The Royal Liverpool Golf Club is one of golf's most storied venues....\\
    Documnet 2 ID: 705
    Document 2 text: Ball dominated amateur golf in Great Britain. He won all the important golf championships as well as the hearts and respect of his country.....\\
    Now I need you to verify the information. Be as objective as possible in your verification. Note you should strictly follow the given format and do not add any extra explanation or information to your output. \\
\rule{\linewidth}{0.4pt}
\textbf{LLM (Searcher):} \{  "query\_id": 0,  
"relevance": [\{"doc\_id": 224,\\
"is\_relevant": true,\\
"is\_relevant\_explanation": "This document mentions John Ball's connection to Hoylake and his influence on golf history, making it relevant to the question." \\
\{"doc\_id": 705,\\
"is\_relevant": true, \\
"is\_relevant\_explanation": "This document also discusses John Ball's influence on golf history and his relationship with Hoylake, making it relevant to the question."\}],\\  "change\_search\_query": false,\\
"change\_search\_query\_explanation": "The current documents are relevant and contain useful information about John Ball's influence on golf at Hoylake.",\\
"new\_search\_query": "",\\ "end\_search": true, \\ "end\_search\_explanation": "The provided documents are sufficient to answer the question about John Ball's influence on golf history at Hoylake."\}
\end{tabular}\\
\hline

\textbf{Coordinator} & 
\begin{tabular}[t]{@{}p{12cm}@{}}
\raggedright
\textbf{User (Searcher):} "only the relevant documents found by searcher"
\rule{\linewidth}{0.4pt}
\textbf{LLM :} \{  "agent": "reasoner",\\ 
"input": \{"question": "How did John Ball influence golf history at Hoylake, and what strategic challenges does the course present regarding out of bounds?", \\
"information": ""John Ball, Jr. was born in Hoylake, Cheshire and grew up playing golf on the Royal Liverpool course. He won several Amateur and Open Championships, including the 1890 Open Championship. He was known for his graceful swing and refusal to use a niblick. In 1878, he competed in his first Open Championship and finished fourth at Prestwick."\}, \\ "aspect": "John Ball's influence on golf history at Hoylake"\}, \\"reason": "I need to analyze the information I've gathered to determine how John Ball influenced golf history at Hoylake."\}
\end{tabular} \\
\hline

\textbf{Reasoner} & 
\begin{tabular}[t]{@{}p{12cm}@{}}
\raggedright
\textbf{User (Coordinator):} Output from Coordinator [Avoiding Repetition]
\rule{\linewidth}{0.4pt}
\textbf{LLM (Reasoner):} \{"analysis": "John Ball, Jr. was born and raised in Hoylake, making him a local figure associated with the Royal Liverpool Golf Club, which is located there.", "He had significant success in golf competitions, winning multiple Amateur and Open Championships, with the most notable being the 1890 Open Championship. This highlights his skill and impact on the sport.", "His participation in the 1878 Open Championship at Prestwick, even before winning the major title, indicates his early prominence in professional golf and his potential influence on the tournament's legacy.", "Ball's graceful swing and refusal to use a niblick (an early type of iron club) reflect his unique style and contributed to his reputation as a skilled player, potentially influencing other golfers' techniques and preferences over time.", "Given his connection to Hoylake and the Royal Liverpool Golf Club, Ball likely played the course extensively, contributing to its historical significance and possibly offering insights or suggestions that shaped the venue's development and reputation."\}
\end{tabular} \\
\hline

\textbf{Coordinator} & 
\begin{tabular}[t]{@{}p{12cm}@{}}
\raggedright
\textbf{User (Reasoner):} "Output from Reasoner [Avoiding Repetition]" 
\rule{\linewidth}{0.4pt}
\textbf{LLM :} \{  "agent": "searcher",\\ 
"input": \{"question": "How did John Ball influence golf history at Hoylake, and what strategic challenges does the course present regarding out of bounds?", \\
"information": "John Ball, Jr. was born and raised in Hoylake, making him a local figure associated with the Royal Liverpool Golf Club, which is located there. He had significant success in golf competitions, winning multiple Amateur and Open Championships, with the most notable being the 1890 Open Championship. His participation in the 1878 Open Championship at Prestwick, even before winning the major title, indicates his early prominence in professional golf and his potential influence on the tournament's legacy. Ball's graceful swing and refusal to use a niblick reflect his unique style and contributed to his reputation as a skilled player, potentially influencing other golfers' techniques and preferences over time. Given his connection to Hoylake and the Royal Liverpool Golf Club, Ball likely played the course extensively, contributing to its historical significance and possibly offering insights or suggestions that shaped the venue's development and reputation."\},\\ "suggestions": ["strategic challenges of Hoylake course", "out of bounds rules in golf", "Royal Liverpool Golf Club history"],\\  "reason": "I need to find more information about the strategic challenges presented by the Hoylake course and the specific rules related to out of bounds to complete the analysis."\}
\end{tabular} \\
\hline

\textbf{Searcher} & 
\begin{tabular}[t]{@{}p{12cm}@{}}
\raggedright
\textbf{User (Coordinator):} Output from Coordinator [Avoiding Repetition]
\rule{\linewidth}{0.4pt}
\textbf{LLM (Searcher):} \{"search\_query": "strategic challenges of Hoylake course",  \\
"search\_query\_explanation": "This search query aims to find information about the strategic challenges presented by the Hoylake course, which is crucial for understanding the second part of the question."\} \\
\rule{\linewidth}{0.4pt}
\textbf{User (Retriever):} This is the information resulted from your search query:\\
    Query ID: 1 \\
    Documnet 1 ID: 399\\
    Document 1 text: The old seventeenth, now the first, known as the Royal hole, was one of Harry Colt\u2019s most famous creations, with the green set right against Stanley Road; sadly, because it was impossible to get spectators round it, the green had to go to get the Open back to Hoylake....\\
    Documnet 2 ID: 829
    Document 2 text: The Open requires mental strength to overcome the obstacles thrown by Mother Nature. Having excellent control of ball flight is essential for any player to be successful when playing in windy, rainy and cold conditions.....\\
    Now I need you to verify the information. Be as objective as possible in your verification. Note you should strictly follow the given format and do not add any extra explanation or information to your output. \\
\rule{\linewidth}{0.4pt}
\textbf{LLM (Searcher):} \{  "query\_id": 1,  
"relevance": [\{"doc\_id": 399,\\
"is\_relevant": true,\\
"is\_relevant\_explanation": "This document discusses the strategic challenges of various golf courses, including Hoylake, particularly focusing on the use of out of bounds lines and bunkers to create strategic elements."\} \\
\{"doc\_id": 829,\\
"is\_relevant": false, \\
"is\_relevant\_explanation": "This document focuses on the mental and physical preparation for playing in difficult weather conditions and does not discuss the strategic challenges of Hoylake or out of bounds rules."\}],\\  "change\_search\_query": false,\\
"change\_search\_query\_explanation": "The current documents are relevant and contain useful information about John Ball's influence on golf at Hoylake.",\\
"new\_search\_query": "",\\ "end\_search": true, \\ "end\_search\_explanation": "The provided documents are sufficient to answer the question about John Ball's influence on golf history at Hoylake."\}
\rule{\linewidth}{0.4pt}
\textbf{User (Retriever):} This is the information resulted from your search query:\\
    Query ID: 2 \\
    Documnet 1 ID: 206\\
    Document 1 text: But it is important to note that, when older courses were built, the penalty for going out of bounds was not necessarily quite as severe as it is today. Back in the day, the use of out of bounds as a strategic hazard was, if not exactly commonplace, far from unknown....\\
    Documnet 2 ID: 399
    Document 2 text: The old seventeenth, now the first, known as the Royal hole, was one of Harry Colt\u2019s most famous creations, with the green set right against Stanley Road; sadly, because it was impossible to get spectators round it, the green had to go to get the Open back to Hoylake... \\
\rule{\linewidth}{0.4pt}
\textbf{LLM (Searcher):} \{  "query\_id": 2,  
"relevance": [\{"doc\_id": 206,\\
"is\_relevant": true,\\
"is\_relevant\_explanation": "This document discusses the strategic use of out of bounds at Hoylake, providing specific examples and insights into how the course utilizes these areas."\} \\
\{"doc\_id": 399,\\
"is\_relevant": true, \\
"is\_relevant\_explanation": "This document also provides detailed information about the strategic use of out of bounds at Hoylake, including examples like the Royal hole and the ninth hole (formerly the seventh)."\}],\\  "change\_search\_query": false,\\
"change\_search\_query\_explanation": "Both documents are relevant and contain useful information about the strategic challenges of Hoylake's course, particularly concerning out of bounds lines.",\\
"new\_search\_query": "",\\ "end\_search": true, \\ "end\_search\_explanation": "The provided documents are sufficient to answer the question about the strategic challenges of the Hoylake course and the role of out of bounds."\}
\end{tabular} \\
\hline

\textbf{Coordinator} & 
\begin{tabular}[t]{@{}p{12cm}@{}}
\raggedright
\textbf{User (Searcher):} "only the relevant documents found by searcher"
\rule{\linewidth}{0.4pt}
\textbf{LLM:} \{"agent": "answerer",\\
"input": \{"question": "How did John Ball influence golf history at Hoylake, and what strategic challenges does the course present regarding out of bounds?",\\
"guidance": "Provide a detailed account of John Ball's influence on golf at Hoylake, highlighting his achievements and unique style. Also, describe the strategic challenges presented by the Hoylake course, particularly focusing on the out of bounds rules and how they affect gameplay."\\
"important\_information": "John Ball's connection to Hoylake, his golfing achievements, and the course's out of bounds rules."\\  "reason": "Based on the gathered information, I will now craft a comprehensive response addressing both aspects of the question."\}
\end{tabular} \\
\hline

\textbf{Answerer} & 
\begin{tabular}[t]{@{}p{12cm}@{}}
\raggedright
\textbf{User (Coordinator):} Output from Coordinator and relevant retrieved documents [Avoiding Repetition]
\rule{\linewidth}{0.4pt}
\textbf{LLM (Answerer):} \{"response": "John Ball significantly influenced golf history at Hoylake, Cheshire, where he grew up playing on the Royal Liverpool Golf Club course. Born in Hoylake, Ball was the son of the Royal Hotel's prosperous owner, which placed him in the heart of the golfing community. His achievements include winning the Open Championship in 1890 and the Amateur Championship in 1888, making him the first English-born player to win both titles in the same year. Ball's unique style was characterized by his graceful swing and his refusal to use a niblick, which he considered 'another bloody spade.' This stubbornness and determination made him a formidable player in match play. Hoylake, known for its storied golfing history, presents strategic challenges, particularly concerning out of bounds rules. The course's design incorporates out of bounds lines as strategic hazards, encouraging players to take risks for better shot lines. For instance, the traditional opening hole features a drive over a turf 'cop,' with the favored strategic line being close to the cop. The use of out of bounds as a strategic element adds complexity to the game, requiring players to balance risk and reward. Overall, John Ball's influence on golf at Hoylake is marked by his exceptional achievements and his impact on the strategic design of the course, particularly through the thoughtful integration of out of bounds rules."\}
\end{tabular} \\
\hline

\textbf{Coordinator} & 
\begin{tabular}[t]{@{}p{12cm}@{}}
\raggedright
\textbf{User (Answerer):} Output from Answerer [Avoiding Repetition] \rule{\linewidth}{0.4pt}
\textbf{LLM:}  \{"agent": "validator", "input": \{ "question":"How did John Ball influence golf history at Hoylake, and what strategic
challenges does the course present regarding out of bounds?"\\ "information": "John Ball significantly influenced golf history at Hoylake, Cheshire, where he grew up playing on the Royal Liverpool Golf Club course. Born in Hoylake, Ball was the son of the Royal Hotel's prosperous owner, which placed him in the heart of the golfing community. His achievements include winning the Open Championship in 1890 and the Amateur Championship in 1888, making him the first English-born player to win both titles in the same year. Ball's unique style was characterized by his graceful swing and his refusal to use a niblick, which he considered 'another bloody spade.' This stubbornness and determination made him a formidable player in match play. Hoylake, known for its storied golfing history, presents strategic challenges, particularly concerning out of bounds rules. The course's design incorporates out of bounds lines as strategic hazards, encouraging players to take risks for better shot lines. For instance, the traditional opening hole features a drive over a turf 'cop,' with the favored strategic line being close to the cop. The use of out of bounds as a strategic element adds complexity to the game, requiring players to balance risk and reward. Overall, John Ball's influence on golf at Hoylake is marked by his exceptional achievements and his impact on the strategic design of the course, particularly through the thoughtful integration of out of bounds rules.",\\  "response": "John Ball significantly influenced golf history at Hoylake, Cheshire, where he grew up playing on the Royal Liverpool Golf Club course. Born in Hoylake, Ball was the son of the Royal Hotel's prosperous owner, which placed him in the heart of the golfing community. His achievements include winning the Open Championship in 1890 and the Amateur Championship in 1888, making him the first English-born player to win both titles in the same year. Ball's unique style was characterized by his graceful swing and his refusal to use a niblick, which he considered 'another bloody spade.' This stubbornness and determination made him a formidable player in match play. Hoylake, known for its storied golfing history, presents strategic challenges, particularly concerning out of bounds rules. The course's design incorporates out of bounds lines as strategic hazards, encouraging players to take risks for better shot lines. For instance, the traditional opening hole features a drive over a turf 'cop,' with the favored strategic line being close to the cop. The use of out of bounds as a strategic element adds complexity to the game, requiring players to balance risk and reward. Overall, John Ball's influence on golf at Hoylake is marked by his exceptional achievements and his impact on the strategic design of the course, particularly through the thoughtful integration of out of bounds rules."\},\\ "reason": "I need to validate if the response accurately addresses the question and is grounded in the provided information."\}
\end{tabular} \\
\hline

\textbf{Validator} & 
\begin{tabular}[t]{@{}p{12cm}@{}}
\raggedright
\textbf{User (Coordinator):} Output from Coordinator [Avoiding Repetition]
\rule{\linewidth}{0.4pt}
\textbf{LLM (Validator):} \{"extracted\_criteria": [\{"criteria": "detailed", "criteria\_explanation": ""The question asks about John Ball's influence on golf history at Hoylake and the strategic challenges presented by the course. A detailed response would cover these aspects comprehensively.", "is\_response\_valid\": true, "is\_response\_valid\_feedback": "The response is detailed and covers both the historical influence of John Ball and the strategic challenges presented by the course."\}], \\
"is\_groundedly\_supported": true, \\ "is\_groundedly\_supported\_feedback": "The response is well-grounded in the provided information and includes specific examples such as the Open Championship wins and the strategic use of out of bounds lines.", \\
"is\_correctly\_answered\": true,\\
"is\_correctly\_answered\_feedback\": \"The response correctly answers the question by detailing John Ball's influence and explaining the strategic challenges presented by Hoylake."\}
\}
\end{tabular}\\
\hline

\textbf{Coordinator} & 
\begin{tabular}[t]{@{}p{12cm}@{}}
\raggedright
\textbf{User (Validator):} Output from Validator [Avoiding Repetition]
\rule{\linewidth}{0.4pt}
\textbf{LLM:} \{"agent":"finisher", "input":{"finished":true}\}
\end{tabular}\\
\hline

\end{longtable}

\twocolumn